\pdfoutput=1
\PassOptionsToPackage{table,xcdraw}{xcolor}

\documentclass[11pt]{article}

\usepackage{acl}

\usepackage{times}
\usepackage{latexsym}

\usepackage[T1]{fontenc}
\usepackage[utf8]{inputenc}

\usepackage{microtype}

\usepackage{inconsolata}

\usepackage{graphicx}

\usepackage{mathtools}
\usepackage{amsmath}
\usepackage{amssymb}
\usepackage{tabularx}
\usepackage{adjustbox}
\usepackage{tcolorbox}
\usepackage{float}
\usepackage{pifont}
\usepackage{comment}
\usepackage[normalem]{ulem} % for underlining
\useunder{\uline}{\ul}{}
\usepackage{fontawesome5}
\usepackage{booktabs}    % For improved table lines
\usepackage{array}       % For better column control
\usepackage{arydshln} % Draw dash-lines
\usepackage{multirow} % For multirow feature
\usepackage[group-separator={,}]{siunitx} % For numbers
\definecolor{darkgreen}{rgb}{0.0, 0.6, 0.0}
\definecolor{customRed}{RGB}{230,62,48}    %% A bright red
\definecolor{lightgray}{gray}{0.98}
\definecolor{mildgray}{gray}{0.3} % Define mild grey for the frame
\definecolor{darkred}{rgb}{0.7,0,0}
\definecolor{lightblue}{rgb}{0.9, 0.95, 1}
\definecolor{darkblue}{rgb}{0.1, 0.3, 0.6}
\definecolor{midblue}{rgb}{0.2, 0.6, 0.6}

\newcommand{\sref}[1]{\S\ref{#1}}

\newcommand{\ThickVLine}{\rule{1pt}{2.25ex}} % Change 1pt to desired thickness

\newcommand{\dingcheck}{\ding{51}}
\newcommand{\dingcross}{\ding{55}}

\title{LFOSum: Summarizing Long-form Opinions with Large Language Models}

\author{Mir Tafseer Nayeem \\
  University of Alberta \\
  \texttt{mnayeem@ualberta.ca} \\\And
  Davood Rafiei \\
  University of Alberta \\
  \texttt{drafiei@ualberta.ca} \\}

\begin{document}
\maketitle
\begin{abstract}
Online reviews play a pivotal role in influencing consumer decisions across various domains, from purchasing products to selecting hotels or restaurants. However, the sheer volume of reviews—often containing repetitive or irrelevant content—leads to information overload, making it challenging for users to extract meaningful insights. 
Traditional opinion summarization models face challenges in handling long inputs and large volumes of reviews, while newer Large Language Model (LLM) approaches often fail to generate accurate and faithful summaries.
To address those challenges, this paper introduces (1) a new dataset of long-form user reviews, each entity comprising over a thousand reviews, (2) two training-free LLM-based summarization approaches that scale to long inputs, and (3) automatic evaluation metrics.
Our dataset of user reviews is paired with in-depth and unbiased critical summaries by domain experts, serving as a reference for evaluation. Additionally, our novel reference-free evaluation metrics provide a more granular, context-sensitive assessment of summary faithfulness. We benchmark several open-source and closed-source LLMs using our methods. Our evaluation reveals that LLMs still face challenges in balancing sentiment and format adherence in long-form summaries, though open-source models can narrow the gap when relevant information is retrieved in a focused manner\footnote{We will make our dataset, code, and outputs publicly available at \href{https://github.com/tafseer-nayeem/LFOSum}{LFOSum}.}.
\end{abstract}

\section{Introduction}
\label{sec:introduction}

Online opinions play a critical role in shaping consumer decisions about what products to buy, where to stay, where to eat, and even which books to read. A recent survey found that approximately \num{98}\% of online customers read reviews before making a purchase decision \citep{power-reviews-article}. These reviews reflect user opinions, providing valuable insights that help set realistic expectations and reveal key details about products and services. However, popular products often accumulate hundreds or even thousands of reviews, many of which contain uninformative content, such as irrelevant personal anecdotes, making them overwhelming to sift through. This leads to information overload \citep{10.1086/208982}, where the sheer volume of reviews discourages consumers, sometimes disregarding the reviews at all \citep{Soto-Acosta:2014:1468-4527:543}. Market research shows that most customers read fewer than \num{10} reviews before making a purchase \citep{murphy-article}, and this can lead to suboptimal decision-making \citep{kwon2015information}. The sheer volume, variable quality, and limited consumer patience underscore the need for improved review utilization strategies to mitigate information overload and enhance decision-making.

Review summarization has been studied in the literature under the same name~\citep{10.1145/1014052.1014073} and within the broader field of opinion mining and summarization~\citep{INR-011,suhara-etal-2020-opiniondigest}, with the goal of producing a concise and easy-to-read summaries about target entities (e.g., a product, hotel, restaurant, or service).
A well-constructed summary is expected to capture the most common or popular viewpoints while omitting unnecessary or irrelevant information \citep{ganesan-etal-2010-opinosis, hosking2024hierarchical}. 
A key challenge is the scarcity of annotated datasets that pair reviews with summaries. Most review platforms do not provide summaries, and creating them would require costly human annotation,
unlike news summarization datasets \citep{NIPS2015-afdec700, see-etal-2017-get, narayan-etal-2018-dont}, where summaries are often included in the source documents. To address this, existing studies have leveraged self-supervised approaches, generating synthetic pairs from review corpora \citep{amplayo-lapata-2020-unsupervised, elsahar-etal-2021-self}, typically by designating one review as a pseudo-summary of others. However, most of these datasets are limited to a maximum of \num{10} reviews \citep{angelidis-lapata-2018-summarizing, pmlr-v97-chu19b, brazinskas-etal-2020-shot}, with only a few extending to hundreds \cite{angelidis-etal-2021-extractive, brazinskas-etal-2021-learning}, while real-world entities often accumulate thousands of reviews. Our work aims to scale review summarization to accommodate larger volumes of reviews.

An effective opinion summarization model should possess several desirable properties to address the challenges associated with large-scale review summarization \cite{10.1145/3477495.3532676}.
First, it should offer \textbf{control mechanisms}~\citep{amplayo-etal-2021-aspect, li-etal-2023-aspect}, enabling users to customize the summaries to their specific needs. Second, the model must be \textbf{scalable}, capable of processing thousands of user opinions while efficiently extracting essential information \citep{hosking-etal-2023-attributable}. Lastly, the generated summaries must be \textbf{faithful} to the input texts, accurately representing their content while minimizing the risk of hallucination~\citep{maynez-etal-2020-faithfulness, tang-etal-2023-understanding}.

In this paper, we explore three control mechanisms for opinion summarization: \textbf{(1)} query control, \textbf{(2)} sentiment control, and \textbf{(3)} length control. With query control, users can specify preferences such as \emph{‘ocean view’} or proximity to a \emph{‘metro station.’} Sentiment control enables structuring summaries into sections like \texttt{‘PROS’} and \texttt{‘CONS’}, while length control allows users to dictate the length of the generated summaries.
To handle large volumes of reviews, we examine two scalable approaches: Retrieval-Augmented Generation (RAG) and long-context Large Language Models (LLMs) \citep{lee2024longcontextlanguagemodelssubsume}, both of which show promise \citep{li2024retrievalaugmentedgenerationlongcontext}. Evaluating faithfulness in long-form summarization poses a unique challenge \citep{siledar-etal-2024-one}, as modern models often suffer from hallucinations \cite{maynez-etal-2020-faithfulness, tang-etal-2023-understanding}. Traditional metrics like \texttt{RAGAs} \cite{es-etal-2024-ragas} and \texttt{RAGChecker} \cite{ru2024ragcheckerfinegrainedframeworkdiagnosing} are typically designed for factual tasks such as question answering or knowledge-based generation, where sentiment and opinions are secondary concerns.
To better align generated summaries with input texts, we treat both as sets of triplets and develop a scheme to quantify their alignment. This approach offers a reference-free evaluation metric tailored to sentiment-rich domains, such as product and service reviews, where opinion and sentiment polarity are crucial.

% Our contributions
Our main contributions are summarized as follows:

\begin{itemize}

    \item We introduce a new dataset of long-form user reviews, where each entity contains over a thousand reviews paired with in-depth, unbiased critical summaries provided by domain experts (\sref{sec:dataset}).  
    %This dataset fills a significant gap in opinion summarization research by enabling the evaluation of models on large-scale, real-world inputs.

    \item We propose training-free methods that utilize RAG and long-context LLMs to address the challenges of long-form opinion summarization. Our approach enables controllable and scalable summarization, providing fine-grained user controls (\sref{sec:methodology}). %such as \texttt{‘PROS’} and \texttt{‘CONS’} divisions, as well as length-based adjustments.

    \item We develop three novel, reference-free automatic evaluation metrics based on Aspect-Opinion-Sentiment (AOS) triplets. These metrics provide a granular and context-sensitive assessment of the faithfulness of generated summaries, particularly in sentiment-rich domains where opinions and sentiment polarity are crucial (\sref{sec:RAG-verification}).
    %, particularly in sentiment-rich domains like product and service reviews.

\end{itemize}

\begin{table*}[t]
\centering
\renewcommand{\arraystretch}{1.05} % Adjusts the row spacing
\resizebox{16cm}{!} %Default width is 16
{
\begin{tabular}{cccccccccc}
\hline
\rowcolor[HTML]{EFEFEF} 
\cellcolor[HTML]{EFEFEF} &
  \cellcolor[HTML]{EFEFEF} &
  \cellcolor[HTML]{EFEFEF} &
  \cellcolor[HTML]{EFEFEF} &
  \cellcolor[HTML]{EFEFEF} &
  \cellcolor[HTML]{EFEFEF} &
  \cellcolor[HTML]{EFEFEF} &
  \cellcolor[HTML]{EFEFEF} &
  \multicolumn{2}{c}{\cellcolor[HTML]{EFEFEF}\textbf{Controls}} \\ \cline{9-10} 
\rowcolor[HTML]{EFEFEF} 
\multirow{-2}{*}{\cellcolor[HTML]{EFEFEF}\textbf{Datasets}} &
  \multirow{-2}{*}{\cellcolor[HTML]{EFEFEF}\textbf{Domain}} &
  \multirow{-2}{*}{\cellcolor[HTML]{EFEFEF}\textbf{\#Entities}} &
  \multirow{-2}{*}{\cellcolor[HTML]{EFEFEF}\textbf{\#Reviews}} &
  \multirow{-2}{*}{\cellcolor[HTML]{EFEFEF}\textbf{\#Sents}} &
  \multirow{-2}{*}{\cellcolor[HTML]{EFEFEF}\textbf{\#Words}} &
  \multirow{-2}{*}{\cellcolor[HTML]{EFEFEF}\textbf{\#Tokens}} &
  \multirow{-2}{*}{\cellcolor[HTML]{EFEFEF}\textbf{\begin{tabular}[c]{@{}c@{}}Book \\ Length?\end{tabular}}} &
  \textbf{Sentiment} &
  \textbf{Length} \\ \hline \hline
\texttt{\textbf{MeanSum}} \citep{pmlr-v97-chu19b}  & Businesses      & 200 & 8 & 41.1 & 542.76 & 561.01 & \dingcross & \dingcross & \dingcross \\
\texttt{\textbf{CopyCat}} \citep{brazinskas-etal-2020-unsupervised} & Products & 60 & 8   & 30.38  & 447.63    & 463.62   & \dingcross & \dingcross & \dingcross \\
\texttt{\textbf{FewSum}} \citep{brazinskas-etal-2020-shot}   & Businesses  & 60  & 8   & 29.85  & 443.6        & 457.05     & \dingcross & \dingcross & \dingcross \\
\texttt{\textbf{OpoSum+}} \citep{amplayo-etal-2021-aspect}      & Products  & 60  & 10   & 71.8   & 1,138.4        & 1,194.0        & \dingcross & \dingcross & \dingcross \\
\texttt{\textbf{SPACE}} \citep{angelidis-etal-2021-extractive}  & Hotels  & 50  & 100   & 910.58   & 16,160.74    & 16,770.18   & \dingcross & \dingcross & \dingcross \\
\texttt{\textbf{AmaSum}} \citep{brazinskas-etal-2021-learning}   & Products  & 3,166  & 322.31   & 1,057.3      & 15,232.26   & 15,614.71    & \dingcross & \dingcheck & \dingcross \\ 
\hdashline
\texttt{\textbf{LFOSum}} (\textit{ours})      & Hotels   & 500  & 1.5K   & 10.5K  & 196K  & 207K  & \dingcheck & \dingcheck & \dingcheck \\\hline
\end{tabular}
}
\caption{Comparison of our LFOSum dataset with existing alternatives, focusing on long-form, book-length inputs (>\num{100}K tokens) and control dimensions. \#Entities refers to the number of entities per dataset, while \#Reviews indicates the average number of reviews per entity. \#Sents represents the average number of sentences per entity, and \#Words and \#Tokens denote the average number of words and tokens (using the \texttt{GPT-4o} tokenizer) per entity.}
\label{tab:dataset-comparison}
\end{table*}

\section{Dataset Construction}
\label{sec:dataset}

We introduce the LFOSum dataset, a collection of long-form user reviews centered around hotel experiences shared online. Rich in detailed descriptions and personal opinions, this dataset is well-suited for opinion summarization tasks. Hotel reviews are particularly valuable due to their in-depth, 
personalized narratives that cover a wide range of user experiences, such as amenities, service quality, and location. Each entity in the dataset contains over a thousand reviews, offering a substantial volume of input texts.

\paragraph{Source Reviews}  The reviews were sourced from TripAdvisor\footnote{\url{https://www.tripadvisor.com}}, a widely-used platform that combines user-generated reviews with online travel booking services. TripAdvisor's reviews, on average, are three times longer than those found on other leading travel platforms \cite{tripadvisor-statistics-article}, making it an ideal resource for exploring the challenges of long-form summarization with book-length inputs (exceeding \num{100}K tokens) \cite{chang2024booookscore}.

\paragraph{Reference Summaries} Annotated datasets that pair summaries with long-form reviews are scarce, largely because such summaries are not readily available on most review platforms and require significant human annotation effort. To address this gap, we utilized Oyster\footnote{\url{https://www.oyster.com}}, a platform specializing in professional hotel reviews. Oyster’s reviews are based on first-hand, in-depth evaluations conducted by expert reviewers, making them a reliable and unbiased source for generating gold-standard summaries. Each review on Oyster is carefully crafted, providing critical assessments that are consistent and trustworthy. The summaries are divided into structured sections, highlighting key aspects of the accommodation, with explicit divisions into \texttt{‘PROS’} and \texttt{‘CONS’}.

\paragraph{Data Pairing and Crawling Process} To construct pairs of input reviews and their corresponding summaries, we identified \num{500} 
%of the most popular 
travel destinations from the Oyster platform. For each entity, we collected the overview section from Oyster, which contains the critical summaries structured into \texttt{‘PROS’} and \texttt{‘CONS’}. Next, we searched for the same entities on TripAdvisor. In some cases, multiple entities had the same name; to disambiguate, we used unique identifiers such as the hotel’s address and postal code. Once we established the correct entity matches, we crawled the relevant user reviews and corresponding summaries to create the dataset (sample in Appendix [Figure \ref{fig:sample-example}]).

\paragraph{Comparison with Existing Datasets} We compare our proposed LFOSum dataset with existing human-referenced datasets used for evaluating opinion summarization models. As shown in Table~\ref{tab:dataset-comparison}, our dataset uniquely features book-length input reviews and supports both sentiment and length control. Although \texttt{AmaSum} \citep{brazinskas-etal-2021-learning} contains more than three times the number of reviews as \texttt{SPACE} \citep{angelidis-etal-2021-extractive}, it has fewer tokens overall due to domain differences as hotel reviews tend to be longer and more detailed. Detailed statistics and preprocessing steps 
%for our LFOSum dataset 
can be found in Appendix (Section \ref{sec:preprocessing} \& Table~\ref{tab:dataset-stats}).

%\paragraph{Automating Critical Summary Generation} One of the primary motivations behind this dataset construction is to automate the generation of Oyster-like critical summaries. Manually creating such summaries is time-consuming and resource-intensive, particularly for less popular destinations that may not receive professional reviews. Our proposed methods aim to automate this process, using both Retrieval-Augmented Generation (RAG) and long-context LLMs to generate high-quality, unbiased critical summaries. This automation allows us to cover the “long tail” of less popular destinations, making the summarization process more scalable and efficient.

\section{Methodology}
\label{sec:methodology}

We propose two scalable, training-free methods to handle large volumes of user reviews effectively. First, the \textbf{Long-form Critic} method directly utilizes long-context LLMs to generate summaries, allowing users to control aspects such as sentiment and length (\sref{sec:long-form-critic}). Second, the \textbf{RAG Framework} combines an extractive-generative approach, managing long sequences by incorporating retrieval augmentation (\sref{sec:RAG-framework}).

\begin{figure*}[t]
    \centering
    \includegraphics[scale = 0.48]{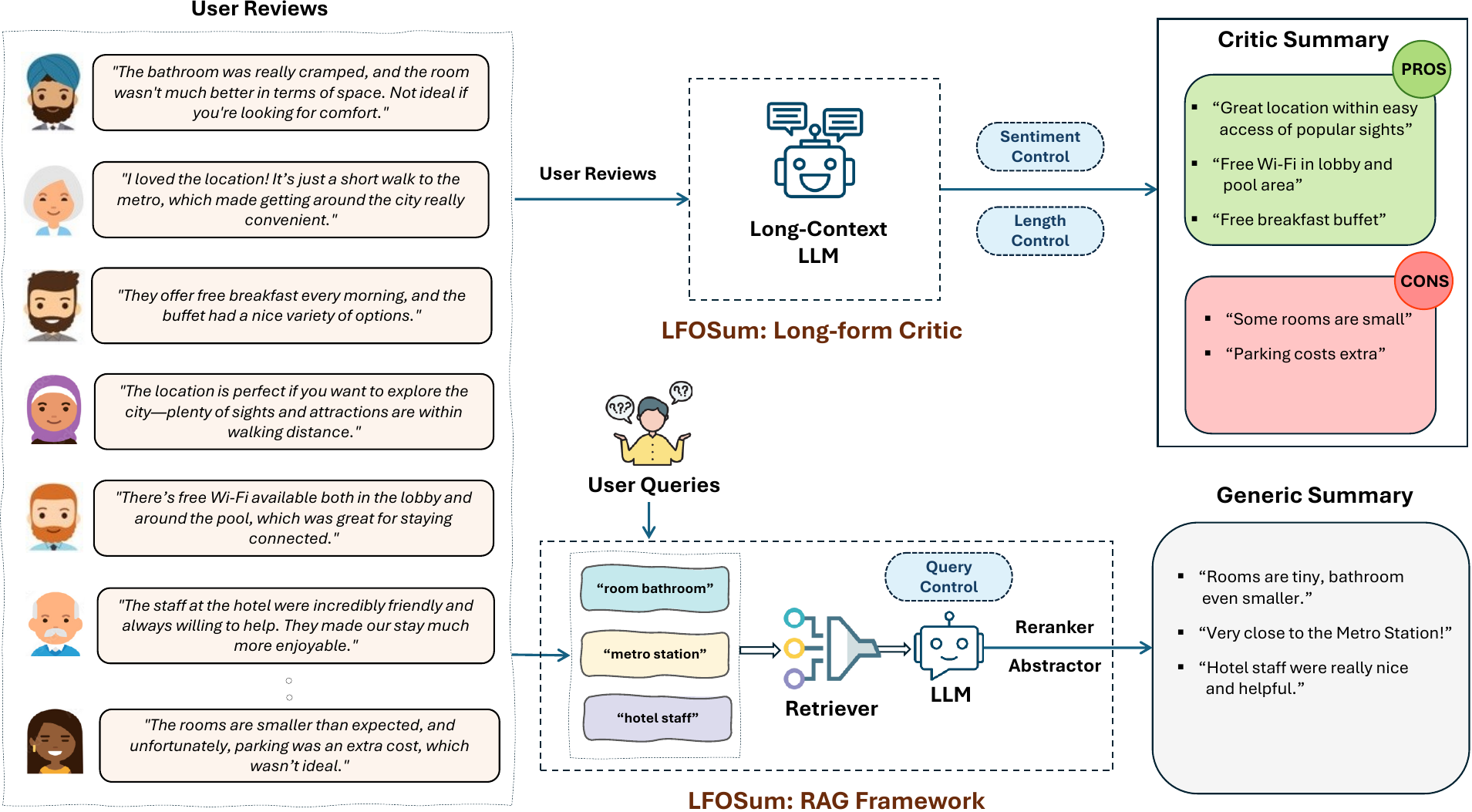}
    \caption{Our LFOSum framework includes two methods: (1) \textbf{Long-form Critic}, which uses long-context LLMs to generate critic summaries with user controls for sentiment and length (\sref{sec:long-form-critic}), and (2) the \textbf{RAG Framework}, which combines retrieval augmentation with LLMs to handle long-form user reviews and produce summaries (\sref{sec:RAG-framework}).}
    %\caption{Overview of our LFOSum Framework.}
    \label{fig:LFOSum-framework}
\end{figure*}

\subsection{LFOSum: Long-form Critic}
\label{sec:long-form-critic}

In this approach, we generate critical summaries consisting of \texttt{‘PROS’} and \texttt{‘CONS’} from the full set of user reviews for a specific entity, presented in a long-form setting. 
%Unlike the retrieval-augmented approach, this method does not filter the input through retrieval techniques; instead, it directly utilizes the complete set of user reviews as input. 
To achieve this, long-context LLMs are employed to process the entire review corpus and generate critical summaries. The LLMs are prompted with a detailed task description, all user reviews for the entity, specific constraints, stylistic exemplars, and are instructed to produce the output in a structured JSON format with separate keys for \texttt{‘PROS’} and \texttt{‘CONS’} (the prompt presented in Figure \ref{fig:critic-prompt} of Appendix). In the basic setting, we do not control the length; the model independently determines the optimal number of \texttt{‘PROS’} and \texttt{‘CONS’} sentences based on the input. The overall process can be formalized as:
\begin{equation}
\text{Critical Summary} = \text{LLM}_{\text{critic}}(R, \mathcal{C}, \mathcal{E}, \mathcal{P})
\end{equation}

Where \( R \) is the set of user reviews, \( \mathcal{C} \) represents task-specific constraints, \( \mathcal{E} \) are stylistic exemplars, and \( \mathcal{P} \) is the task prompt provided to the LLM.

\paragraph{Length Control} In this setting, we introduce a user-centric control mechanism to specify the desired number of \texttt{‘PROS’} and \texttt{‘CONS’} sentences for the critical summary. An additional parameter is included in the LLM prompt to guide the generation length. The number of \texttt{‘PROS’} and \texttt{‘CONS’} is determined based on the ground truth critical summary for the item. By explicitly instructing the LLM with these parameters, we ensure the generated summary aligns with the expected structure and length.

%\paragraph{Format Following}

\paragraph{Sketch $\Rightarrow$ Fetch $\Rightarrow$ Fill (SFF)} To evaluate sentiment and length-controlled summaries, we parse LLM outputs into structured JSON format. However, LLMs sometimes produce incomplete or malformed outputs (some examples in Appendix \ref{sec:parsing-errors}). To address this, we propose the Sketch-Fetch-Fill (SFF) approach for reliable JSON extraction.
%and completion. 

\begin{enumerate}
    \item \textbf{Sketch:} We first define the expected JSON structure, specifying key fields (e.g., `pros' and `cons') to guide reconstruction. %This sketch acts as a template or scaffolding for the JSON structure.
    \item \textbf{Fetch:} Regular expressions are used to extract relevant content from the output, identifying text corresponding to the predefined fields, even with formatting inconsistencies.
    \item \textbf{Fill:} The extracted data is inserted into the predefined structure, correcting common errors (e.g., missing quotes or misplaced commas) to ensure a valid, parsable JSON.
\end{enumerate}

\subsection{LFOSum: RAG Framework}
\label{sec:RAG-framework}

A key component of any RAG framework is the availability of query terms \cite{zhao-etal-2023-retrieving}. In our case, the query terms for an entity are not pre-defined or readily available. To address this, we employ a simple yet effective method to extract query terms from the large input reviews (\sref{sec:query-term-extraction}). These extracted terms are then used to design a combined extractive-generative framework for managing long-form input reviews through retrieval augmentation \cite{10.5555/3495724.3496517}. This approach integrates the attributable and scalable properties of extractive methods (\sref{sec:retrieval}) with the coherence and fluency of LLMs (\sref{sec:reranker-abstractor}). Another advantage of our RAG framework is that it enables the automatic evaluation of generated summaries in manageable units, allowing for a more fine-grained assessment within long-form context (\sref{sec:RAG-verification}).

\subsubsection{Query Term Extraction}
\label{sec:query-term-extraction}

Let $M_e$ denote the language model capturing opinions about an entity $e$, defined as the probability distribution over word sequences.
Under the query likelihood model, entity $e$ is considered \emph{relevant} to query term $q$ if q is \emph{likely generated} by $M_e$. Hence more frequent terms in the reviews of an entity may be treated as important query terms, with the exception of common stop words, and the summarization of an entity may be centered around these important terms.

A related task in the literature is aspect extraction, which can be categorized into two types: \textbf{(1)} Explicit aspects and \textbf{(2)} Implicit aspects~\citep{poria-etal-2014-rule, luo-etal-2018-extra}. Explicit aspects are directly mentioned targets in opinionated sentences, such as \emph{``ocean view''} or \emph{``spa service.''} In contrast, implicit aspects are inherently expressed concepts that can generalize explicit examples; for instance, \emph{``ocean view''} may relate to the broader category of \emph{``location,''} while \emph{``spa service''} falls under \emph{``service.''} In designing our RAG framework, we focus on explicit aspects (referred to as \textbf{\emph{``query terms''}}) due to their repetitive nature in long-form reviews, which facilitates the retrieval of salient sentences covering diverse user concerns. Below, we outline the major components of the query term extraction process:

\paragraph{Candidate Term Extraction \& Ranking} We extract the most frequent unigrams and skip bigrams within a defined window size of 4. This approach captures meaningful multi-word expressions that may not be adjacent but contribute contextually to the overall understanding of the text. To filter out rare or insignificant terms, we apply a frequency threshold, ensuring that only high-frequency, representative terms are retained\footnote{We set the frequency filtering threshold to \num{15}.}. The terms are then ranked based on their frequency values to prioritize those most representative of the input reviews.

\paragraph{Top-K Term Refinement} The extracted query terms are further refined by cross-referencing them with the gold query term list from \cite{pontiki-etal-2015-semeval} for our domain of interest (i.e., Hotel). This helps eliminate frequent but irrelevant terms, such as stop words. To ensure that the final set of terms is diverse and non-redundant, single terms are removed if both of their constituent words appear within a multi-word query. Ultimately, the top-K most relevant query terms are selected for the retrieval step.

\subsubsection{Retrieval}
\label{sec:retrieval}

We divide user reviews into individual sentences and use the Top-K extracted query terms to retrieve relevant sentences as evidence for each query term, which are then provided as input to the LLMs. This approach offers two key advantages: \textbf{(1)} Retrieving sentences based on a diverse set of query terms reduces redundancy in the generated summaries, and \textbf{(2)} it increases information coverage from the user reviews\footnote{Each sentence is assigned to only one query term, and selected sentences are excluded from subsequent selections to prevent overlap.}. The retrieval process is formalized as follows: 
\begin{equation}
S_{Q} = \text{Top-K} \left( \mathcal{R}(Q, D) \right)
\end{equation}

Where \( Q \) is the set of query terms, \( D \) is the collection of review sentences, \( \mathcal{R}(Q, D) \) is the retrieval function, and \( S_{Q} \) represents the Top-K retrieved sentences.

\paragraph{Retrievers} We utilize two types of retrievers: BM25 and Dense retrievers. BM25 is a lexical retriever\footnote{\url{https://github.com/dorianbrown/rank_bm25}} that scores document relevance based on term frequency \cite{10.1561/1500000019}, while Dense retrievers capture deeper contextual meanings through semantic information, ensuring both surface-level lexical matches and nuanced semantic relationships are covered. For the Dense retriever, we employ Sentence Transformers \cite{reimers-gurevych-2019-sentence}, specifically leveraging the checkpoint\footnote{\href{https://huggingface.co/sentence-transformers/all-mpnet-base-v2}{sentence-transformers/all-mpnet-base-v2}} due to its superior performance in semantic search across a wide range of benchmarks.

\subsubsection{LLM as Reranker and Abstractor}
\label{sec:reranker-abstractor}

We utilize the retrieved sentences for each query term as evidence and instruct LLMs to generate summaries. Two variants of summarization approaches are employed: \textbf{(1)} Extractive and \textbf{(2)} Abstractive. In both cases, LLMs are prompted with the retrieved sentences, and the outputs are aligned in a specified JSON format. The general process for both approaches can be formalized as follows:
\begin{equation}
\text{Summary}(Q) = \text{LLM}(Q, S_{Q}, \mathcal{C}, \mathcal{P})
\end{equation}

Where \( Q \) is the query term, \( S_Q \) is the set of Top-K retrieved sentences, \( \mathcal{C} \) represents the constraints, and \( \mathcal{P} \) is the prompt provided to the LLM.

\paragraph{Extractive} In the extractive approach, LLMs are prompted with a task description, constraints, the query term, and a list of Top-K retrieved sentences. The LLM is instructed to rerank the sentences and select the most relevant one, functioning primarily as a reranker. The complete prompt used for this process is shown in Appendix (Figure \ref{fig:reranker-prompt}).

\paragraph{Abstractive} For the abstractive approach, LLMs are prompted with a task description, constraints, the query term, a list of Top-K retrieved sentences, and stylistic exemplars to guide the output in the desired style. The LLM synthesizes a summary based on the retrieved information, effectively acting as an abstractor. The full prompt used for this task is presented in Appendix (Figure \ref{fig:abstractor-prompt}).

\subsubsection{RAG Verification}
\label{sec:RAG-verification}

To evaluate the ability of LLMs to generate summaries that accurately reflect the input evidence, we build upon the work of~\citet{bhaskar-etal-2023-prompted}, who developed desiderata for human evaluation, by introducing automatic evaluation metrics. Our goal is to break down sentences into structured components, allowing for a more granular and fine-grained assessment of factual alignment. We employ Aspect-Opinion-Sentiment (AOS) triplets~\citep{varia-etal-2023-instruction}, using a pre-trained model from~\citet{scaria-etal-2024-instructabsa}, which captures both implicit and explicit aspects (as detailed in \sref{sec:query-term-extraction}). Each triplet decomposes the sentence into three core components:

\begin{itemize}
    \item \textbf{Aspect}: The attribute or feature being discussed (e.g., \emph{``room bathroom''}).
    \item \textbf{Opinion}: The expression or judgment about the aspect (e.g., \emph{``clean''}).
    \item \textbf{Sentiment}: The polarity of the opinion (e.g., \emph{negative}, \emph{neutral}, or \emph{positive}).
\end{itemize}

Given a set of retrieved sentences for each query, and a generated sentence, we evaluate the quality of the generated sentences for the Top-K queries of an entity based on three key metrics:

\begin{itemize} 

    \item \textbf{Aspect Relevance (AR)}: Measures how well the aspect in the generated sentence aligns with the most important and frequent aspects mentioned in the retrieved evidence. This ensures the summary remains on topic and covers critical aspects. 
    \item \textbf{Sentiment Factuality (SF)}: Evaluates for a given aspect whether the sentiment in the generated sentence matches the most frequent sentiment found in the retrieved evidence, ensuring that the sentiment expressed is factually accurate. 
    \item \textbf{Opinion Faithfulness (OF)}: Assesses for a given aspect and sentiment whether the opinion expressed in the generated sentence is consistent with the opinions found in the retrieved evidence, either through direct matching or semantic similarity.
    
\end{itemize}

\paragraph{Aspect Relevance (AR)} For each query, AOS triplets are extracted from both the retrieved and generated sentences. We identify the most frequent aspect from the retrieved evidence and check if it appears in the generated sentence. Aspect Relevance, in this context, is a binary variable, indicating whether the generated sentence remains on-topic by covering the most important aspect. We are interested in the expectation of this variable over generated sentences.

\paragraph{Sentiment Factuality (SF)}  For each aspect, sentiments are extracted from AOS triplets of both the retrieved and generated sentences. Neutral sentiments are excluded as they provide limited insight. For each aspect, the most frequent non-neutral sentiment from the retrieved sentences is identified, and the sentiment in the generated sentence is checked for alignment. Similar to AR, SF is a binary variable, indicating whether the generated sentiment is factually correct. Again, we are interested in the expectation of this variable over generated sentences.

\begin{table*}[t]
\centering
\renewcommand{\arraystretch}{1.3} % Adjusts the row spacing
\resizebox{16cm}{!}  %default is 16
{
\begin{tabular}{cclcccccccccccl}
\hline
\rowcolor[HTML]{EFEFEF} 
\cellcolor[HTML]{EFEFEF} &
  \cellcolor[HTML]{EFEFEF} &
  \multicolumn{1}{c}{\cellcolor[HTML]{EFEFEF}} &
   &
  \multicolumn{3}{c}{\cellcolor[HTML]{EFEFEF}\textbf{PROS Scores}} &
  \textbf{} &
  \multicolumn{3}{c}{\cellcolor[HTML]{EFEFEF}\textbf{CONS Scores}} &
   &
  \multicolumn{3}{c}{\cellcolor[HTML]{EFEFEF}\textbf{Format Following}} \\ \cline{5-7} \cline{9-11} \cline{13-15} 
\rowcolor[HTML]{EFEFEF} 
\multirow{-2}{*}{\cellcolor[HTML]{EFEFEF}\textbf{Models}} &
  \multirow{-2}{*}{\cellcolor[HTML]{EFEFEF}\textbf{\begin{tabular}[c]{@{}c@{}}Context\\ Length\end{tabular}}} &
  \multicolumn{1}{c}{\multirow{-2}{*}{\cellcolor[HTML]{EFEFEF}\textbf{Settings}}} &
   &
  \textbf{R1} &
  \textbf{RL} &
  \textbf{BERTScore} &
  \textbf{} &
  \textbf{R1} &
  \textbf{RL} &
  \textbf{BERTScore} &
   &
  \textbf{JSON Parsing} &
  \multicolumn{2}{c}{\cellcolor[HTML]{EFEFEF} \textbf{SFF} (\textit{ours})} \\ \hline \hline
 &
   &
  \textbf{GPT-4o-mini} &
   &
  29.87 &
  17.15 &
  65.40 &
   &
  15.68 &
  9.66 &
  57.06 &
   &
  500 / 500 &
  \multicolumn{2}{c}{500 / 500} \\
\multirow{-2}{*}{\includegraphics[width=0.65cm]{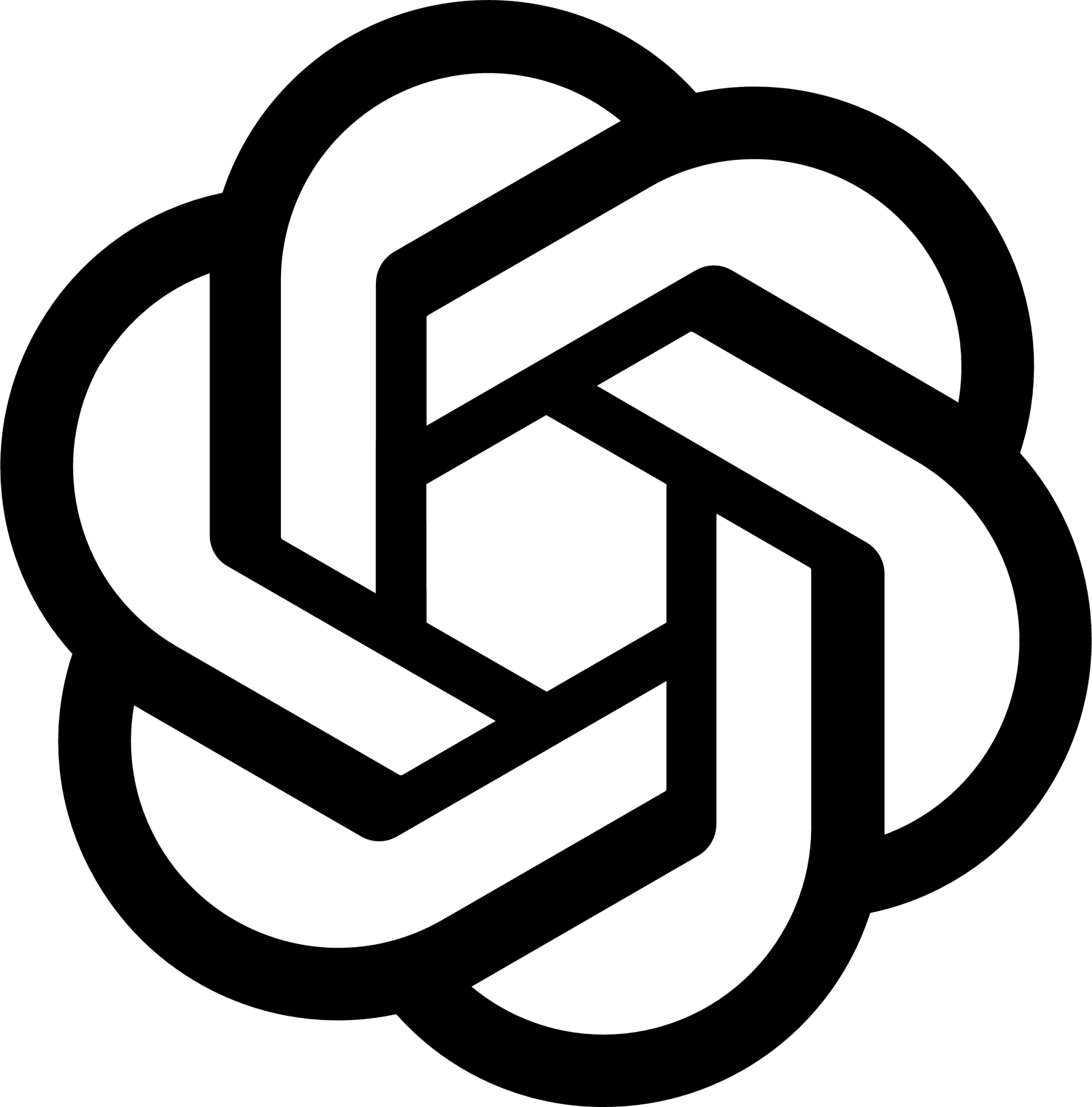}} &
  \multirow{-2}{*}{\num{128}K} &
  \cellcolor[HTML]{F1F1F1} \enspace \ThickVLine \textbf{--} w/ Length Control \ \faEdit &
  \cellcolor[HTML]{F1F1F1} &
  \cellcolor[HTML]{F1F1F1}\textbf{30.99} &
  \cellcolor[HTML]{F1F1F1}\textbf{17.75} &
  \cellcolor[HTML]{F1F1F1}\textbf{66.24} &
  \cellcolor[HTML]{F1F1F1} &
  \cellcolor[HTML]{F1F1F1}18.39 &
  \cellcolor[HTML]{F1F1F1}12.03 &
  \cellcolor[HTML]{F1F1F1}60.31 &
  \cellcolor[HTML]{F1F1F1} &
  \cellcolor[HTML]{F1F1F1}500 / 500 &
  \multicolumn{2}{c}{\cellcolor[HTML]{F1F1F1}500 / 500} \\ \hline
 &
   &
  \textbf{Claude-3-Haiku} &
   &
  30.67 &
  {\ul 18.08} &
  {\ul 66.74} &
   &
  {\ul 20.19} &
  {\ul 12.91} &
  {\ul 60.85} &
   &
  457 / 500 &
  \multicolumn{2}{c}{500 / 500} \\
\multirow{-2}{*}{\includegraphics[width=0.65cm]{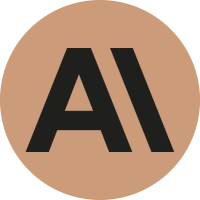}} &
  \multirow{-2}{*}{\num{200}K} &
  \cellcolor[HTML]{F1F1F1} \enspace \ThickVLine \textbf{--} w/ Length Control \ \faEdit &
  \cellcolor[HTML]{F1F1F1} &
  \cellcolor[HTML]{F1F1F1}30.50 &
  \cellcolor[HTML]{F1F1F1}17.41 &
  \cellcolor[HTML]{F1F1F1}66.07 &
  \cellcolor[HTML]{F1F1F1} &
  \cellcolor[HTML]{F1F1F1}19.67 &
  \cellcolor[HTML]{F1F1F1}12.24 &
  \cellcolor[HTML]{F1F1F1}61.09 &
  \cellcolor[HTML]{F1F1F1} &
  \cellcolor[HTML]{F1F1F1}440 / 500 &
  \multicolumn{2}{c}{\cellcolor[HTML]{F1F1F1} 500 / 500} \\ \hline
 &
   &
  \textbf{Gemini-1.5-Flash} &
   &
  {\ul 31.48} &
  17.80 &
  65.84 &
   &
  17.77 &
  11.26 &
  58.11 &
   &
  343 / 500 &
  \multicolumn{2}{c}{372 / 372} \\ 
\multirow{-2}{*}{\includegraphics[width=0.95cm]{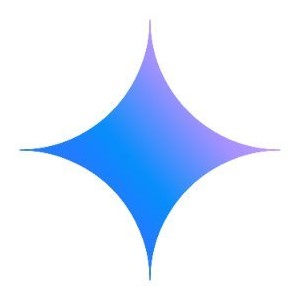}} &
  \multirow{-2}{*}{\num{1}M} &
  \cellcolor[HTML]{F1F1F1} \enspace \ThickVLine \textbf{--} w/ Length Control \ \faEdit &
  \cellcolor[HTML]{F1F1F1} &
  \cellcolor[HTML]{F1F1F1}30.34 &
  \cellcolor[HTML]{F1F1F1}17.65 &
  \cellcolor[HTML]{F1F1F1}65.71 &
  \cellcolor[HTML]{F1F1F1} &
  \cellcolor[HTML]{F1F1F1}\textbf{18.87} &
  \cellcolor[HTML]{F1F1F1}\textbf{12.92} &
  \cellcolor[HTML]{F1F1F1}\textbf{61.24} &
  \cellcolor[HTML]{F1F1F1} &
  \cellcolor[HTML]{F1F1F1}489 / 500 &
  \multicolumn{2}{c}{\cellcolor[HTML]{F1F1F1}495 / 495} \\ \hline
\end{tabular}
}
\caption{Evaluation results of our LFOSum: \textbf{Long-form Critic} method. PROS refers to positive summaries and CONS refers to negative summaries. Best scores for the Length Control setting are marked in \textbf{bold}, while the highest results in the basic setting are {\ul underlined}. ``JSON Parsing'' shows the number of samples successfully parsed directly, and ``SFF (\textit{ours})'' indicates samples recovered using our SFF method from the valid summaries.}
\label{tab:long-context-evals}
\end{table*}

\paragraph{Opinion Faithfulness (OF)} For each aspect and sentiment, opinions are extracted from AOS triplets of both retrieved and generated sentences. A direct opinion match is assigned a score of \num{1}, while indirect matches are evaluated using a semantic similarity function (e.g., cosine similarity), which returns a value between \num{0} and \num{1}. This allows for semantically similar opinions (e.g., \emph{``beautiful''} and \emph{``stunning''}) to be considered faithful. Therefore, the opinion faithfulness for a given aspect and sentiment is represented as a random variable ranging from \num{0} to \num{1}, and we report its expected value over generated sentences.

\section{Evaluation}

In this section, we evaluate the performance of our two proposed approaches: \textbf{(1)} the Long-form Critic (\sref{sec:eval-long-form}) and \textbf{(2)} the RAG Framework (\sref{sec:eval-RAG-framework}). We assess these methods using a variety of open-source and closed-source models, comparing their performance on standard and newly proposed evaluation metrics. The experimental setup is detailed in the Appendix \ref{sec:experimental-setup}, generated summaries in Table \ref{tab:output-summaries-critic-1}, Table \ref{tab:output-summaries-critic-2} \& Table \ref{tab:output-summaries-RAG}, and  the related works are covered in Appendix \ref{sec:related-work}.

\paragraph{Automatic Evaluation} We use F1 scores of ROUGE (R1 and RL) \cite{lin-2004-rouge} and BERTScore \cite{Zhang*2020BERTScore:}, following \cite{bhaskar-etal-2023-prompted}. Although ROUGE scores have been shown to be less reliable for generic opinion summarization tasks \cite{tay-etal-2019-red, shen2023opinsummeval}, we report them for consistency with recent studies \cite{bhaskar-etal-2023-prompted, lei-etal-2024-polarity, siledar-etal-2024-product, hosking2024hierarchical}, to benchmark our dataset and methods in long-form settings, and to contribute to discussions on automatic evaluation methods for long-form opinion summarization (\sref{sec:discussions}). Additionally, we use our proposed evaluation metrics to assess the faithfulness of the LLM-generated summaries.

\subsection{Evaluating Long-form Critic}
\label{sec:eval-long-form}

We evaluate the ability of several LLMs to generate critical summaries divided into \texttt{‘PROS’} and \texttt{‘CONS’}. For this purpose, we utilized long-context LLMs, providing the full set of user reviews as input. 
%The models were instructed to generate summaries in a fixed JSON format, as detailed in Appendix \ref{sec:parsing-errors}. 
We experimented with several closed-source models, including \texttt{GPT-4o-mini}\footnote{\href{https://platform.openai.com/docs/models/gpt-4o}{OpenAI (GPT-4o-mini Model)}}, \texttt{Claude-3-Haiku}\footnote{\href{https://www.anthropic.com/news/claude-3-haiku}{Anthropic (Claude-3-Haiku Model)}}, and \texttt{Gemini-1.5-Flash}\footnote{\href{https://deepmind.google/technologies/gemini/flash/}{Google (Gemini-1.5-Flash Model)}}, alongside open-source models such as \texttt{Llama-3.2-3B-Instruct}~\citep{dubey2024llama3herdmodels} and \texttt{Phi-3.5-mini-instruct}~\citep{abdin2024phi3}, each with varying context lengths. 

However, we encountered significant challenges with open-source models. As highlighted in \cite{xia-etal-2024-fofo}, these models frequently failed to adhere to the expected output format, often producing non-parsable JSON outputs, even when employing our SFF method for parsing (\sref{sec:long-form-critic}). %Given these limitations, we opted not to pursue further experiments with the open-source models for this task. 
We directly parsed the expected JSON outputs from the LLMs, and in cases of errors (detailed in the Appendix \ref{sec:parsing-errors}), we attempted to automatically recover them using our SFF method (\sref{sec:long-form-critic}). If the context length exceeded the model's limit, we truncated the older reviews, prioritizing more recent ones based on posting dates. Summaries were considered valid only if both the "pros" and "cons" sections were not empty, and any invalid summaries were excluded from the evaluation.

\paragraph{Results \& Analysis} As shown in Table \ref{tab:long-context-evals}, Claude-3-Haiku produces the best summaries in the basic setting for both \texttt{‘PROS’} and \texttt{‘CONS’}. However, across all models, \texttt{‘CONS’} performance is generally weaker, likely because negative reviews are less frequent compared to positive ones \cite{venkatesakumar2021distribution}, making it harder for the models to capture \emph{``needle-in-a-haystack''} information within long-form inputs \cite{laban2024summaryhaystack}. In the length control setting, GPT-4o-mini excels in \texttt{‘PROS’}, while Gemini-1.5-Flash performs better in \texttt{‘CONS’}, likely due to its larger context window. Claude-3-Haiku struggles with length adherence, as noted in Appendix (Table \ref{tab:length-control-results}). Gemini-1.5-Flash generated \num{372} out of \num{500} valid summaries, with the remaining invalid due to empty fields, elaborated more in \sref{sec:discussions}. These results highlight the challenge of balancing sentiment and format adherence in long-form summaries.

\subsection{Evaluating RAG Framework}
\label{sec:eval-RAG-framework}
We evaluate our RAG Framework using both open-source and closed-source models. A maximum of \num{15} top query terms (K=\num{15}) are selected for the retrievers, and for each query term, we experiment with retrieving \num{10} and \num{20} sentences. For both summary variants—(1) Extractive and (2) Abstractive—the system-generated summary is created by merging the sentences for each query term, as detailed in \sref{sec:reranker-abstractor}. The \texttt{‘PROS’} and \texttt{‘CONS’} from the gold summaries are merged to form a generic reference summary, following the standard opinion summarization evaluation protocol without sentiment control~\citep{bhaskar-etal-2023-prompted}.

\paragraph{Baselines} For the BM25 and Dense baselines, we select the top sentence retrieved for each of the K query terms to form the summary. For the random baseline, K sentences are randomly selected from the input reviews for each entity. As an upper-bound baseline, the Oracle selects the sentence with the highest ROUGE-L (RL) score for each gold summary sentence, providing an approximate upper limit for performance.

\begin{table}[t]
\centering
\renewcommand{\arraystretch}{1} % Adjusts the row spacing
\resizebox{7.75cm}{!} 
{
\begin{tabular}{lccc}
\hline
\rowcolor[HTML]{EFEFEF} 
\multicolumn{1}{c}{\cellcolor[HTML]{EFEFEF}\textbf{Models (K=20)}} & \textbf{R1} & \textbf{RL} & \textbf{BERTScore} \\ \hline \hline
\textbf{Random}             & 16.35          & 8.45         & 47.42         \\
\textbf{Oracle}       & 51.91          & 44.79         & 67.61         \\
\hdashline
\textbf{BM25}               & 20.93      & 10.34     & 53.15     \\
\textbf{Dense}              & 20.58      & 10.56     & 54.26     \\ \hline
\rowcolor[HTML]{EFEFEF} 
\multicolumn{4}{c}{\cellcolor[HTML]{EFEFEF}\textbf{Extractive}}  \\ \hline 
\textbf{BM25}               &            &           &           \\
\enspace \enspace \ThickVLine \textbf{--} \textbf{\texttt{Mistral-7B}}         & 20.45 (\textcolor{customRed}{-\num{0.48}})      & 10.19 (\textcolor{customRed}{-\num{0.15}})      & 53.46 (\textcolor{darkgreen}{+\num{0.31}})     \\
\enspace \enspace \ThickVLine \textbf{--} \textbf{\texttt{Llama-3-8B}}         & 21.48 (\textcolor{darkgreen}{+\num{0.55}})      & 10.60 (\textcolor{darkgreen}{+\num{0.26}})      & 54.64 (\textcolor{darkgreen}{+\num{1.49}})     \\
\enspace \enspace \ThickVLine \textbf{--} \textbf{\texttt{Gemma-2-9B}}         & 21.61 (\textcolor{darkgreen}{+\num{0.68}})       & 10.87 (\textcolor{darkgreen}{+\num{0.53}})     & 54.89 (\textcolor{darkgreen}{+\num{1.74}})     \\
\enspace \enspace \ThickVLine \textbf{--} \textbf{\texttt{Claude-3-Haiku}}            & \textbf{21.69} (\textcolor{darkgreen}{+\num{0.76}})           & 10.90 (\textcolor{darkgreen}{+\num{0.56}})         & 55.03 (\textcolor{darkgreen}{+\num{1.88}})        \\ 
\enspace \enspace \ThickVLine \textbf{--} \textbf{\texttt{GPT-4o-mini}}            & 21.67 (\textcolor{darkgreen}{+\num{0.74}})          & \textbf{11.02} (\textcolor{darkgreen}{+\num{0.68}})         & \textbf{55.20} (\textcolor{darkgreen}{+\num{2.05}})        \\ \hline
\textbf{Dense}              &            &           &           \\
\enspace \enspace \ThickVLine \textbf{--} \textbf{\texttt{Mistral-7B}}         & 21.17 (\textcolor{darkgreen}{+\num{0.59}})     & 10.61 (\textcolor{darkgreen}{+\num{0.05}})     & 54.43 (\textcolor{darkgreen}{+\num{0.17}})    \\
\enspace \enspace \ThickVLine \textbf{--} \textbf{\texttt{Llama-3-8B}}         & 21.99 (\textcolor{darkgreen}{+\num{1.41}})      & 11.0 (\textcolor{darkgreen}{+\num{0.44}})     & 55.41 (\textcolor{darkgreen}{+\num{1.15}})    \\
\enspace \enspace \ThickVLine \textbf{--} \textbf{\texttt{Gemma-2-9B}}         & 22.28 (\textcolor{darkgreen}{+\num{1.70}})     & 11.15 (\textcolor{darkgreen}{+\num{0.59}})      & 55.50 (\textcolor{darkgreen}{+\num{1.24}})      \\
\enspace \enspace \ThickVLine \textbf{--} \textbf{\texttt{Claude-3-Haiku}}            & 22.06 (\textcolor{darkgreen}{+\num{1.48}})         & 11.16 (\textcolor{darkgreen}{+\num{0.60}})         & 55.56 (\textcolor{darkgreen}{+\num{1.30}})        \\ 
\enspace \enspace \ThickVLine \textbf{--} \textbf{\texttt{GPT-4o-mini}}            & \textbf{22.68} (\textcolor{darkgreen}{+\num{2.10}})         & \textbf{11.41} (\textcolor{darkgreen}{+\num{0.85}})        & \textbf{55.98} (\textcolor{darkgreen}{+\num{1.72}})         \\ \hline
\rowcolor[HTML]{EFEFEF} 
\multicolumn{4}{c}{\cellcolor[HTML]{EFEFEF}\textbf{Abstractive}} \\ \hline 
\textbf{BM25}               &            &           &           \\
\enspace \enspace \ThickVLine \textbf{--} \textbf{\texttt{Mistral-7B}}         & 23.28 (\textcolor{darkgreen}{+\num{2.35}})      & 11.85 (\textcolor{darkgreen}{+\num{1.51}})     & 55.67 (\textcolor{darkgreen}{+\num{2.52}})     \\
\enspace \enspace \ThickVLine \textbf{--} \textbf{\texttt{Llama-3-8B}}         & \textbf{25.52} (\textcolor{darkgreen}{+\num{4.59}})      & \textbf{13.53} (\textcolor{darkgreen}{+\num{3.19}})     & \textbf{56.57} (\textcolor{darkgreen}{+\num{3.42}})     \\
\enspace \enspace \ThickVLine \textbf{--} \textbf{\texttt{Gemma-2-9B}}         & 22.50 (\textcolor{darkgreen}{+\num{1.57}})       & 11.65 (\textcolor{darkgreen}{+\num{1.31}})     & 54.76 (\textcolor{darkgreen}{+\num{1.61}})     \\
\enspace \enspace \ThickVLine \textbf{--} \textbf{\texttt{Claude-3-Haiku}}            & 23.39 (\textcolor{darkgreen}{+\num{2.46}})         & 12.75 (\textcolor{darkgreen}{+\num{2.41}})        & 55.81 (\textcolor{darkgreen}{+\num{2.66}})         \\ 
\enspace \enspace \ThickVLine \textbf{--} \textbf{\texttt{GPT-4o-mini}}            & 24.13 (\textcolor{darkgreen}{+\num{3.20}})         & 12.76 (\textcolor{darkgreen}{+\num{2.42}})        & 55.27 (\textcolor{darkgreen}{+\num{2.12}})         \\ \hline
\textbf{Dense}              &            &           &           \\
\enspace \enspace \ThickVLine \textbf{--} \textbf{\texttt{Mistral-7B}}         & 23.21 (\textcolor{darkgreen}{+\num{2.63}})      & 12.24 (\textcolor{darkgreen}{+\num{1.68}})     & 55.86 (\textcolor{darkgreen}{+\num{1.60}})     \\
\enspace \enspace \ThickVLine \textbf{--} \textbf{\texttt{Llama-3-8B}}         & \textbf{24.62} (\textcolor{darkgreen}{+\num{4.04}})      & \textbf{13.48} (\textcolor{darkgreen}{+\num{2.92}})     & \textbf{56.19} (\textcolor{darkgreen}{+\num{1.93}})     \\
\enspace \enspace \ThickVLine \textbf{--} \textbf{\texttt{Gemma-2-9B}}         & 22.84 (\textcolor{darkgreen}{+\num{2.26}})      & 12.06 (\textcolor{darkgreen}{+\num{1.50}})     & 55.13 (\textcolor{darkgreen}{+\num{0.87}})     \\
\enspace \enspace \ThickVLine \textbf{--} \textbf{\texttt{Claude-3-Haiku}}            & 23.06 (\textcolor{darkgreen}{+\num{2.48}})         & 12.83 (\textcolor{darkgreen}{+\num{2.27}})        & 55.56 (\textcolor{darkgreen}{+\num{1.30}})         \\ 
\enspace \enspace \ThickVLine \textbf{--} \textbf{\texttt{GPT-4o-mini}}            & 24.16 (\textcolor{darkgreen}{+\num{3.58}})          & 12.88 (\textcolor{darkgreen}{+\num{2.32}})        & 55.59 (\textcolor{darkgreen}{+\num{1.33}})        \\ \hline
\end{tabular}
}
\caption{Evaluation results of our LFOSum: RAG Framework with K=\num{20}, where K is the number of retrieved sentences. The best results compared to their respective baseline models are marked in \textbf{bold}, and $\Delta$ gains are shown in round brackets and highlighted in \textcolor{darkgreen}{green} for improvements and \textcolor{customRed}{red} for declines.}
\label{tab:RAG-20}
\end{table}

\begin{table}[t]
\centering
\renewcommand{\arraystretch}{1} % Adjusts the row spacing
\resizebox{7.75cm}{!} 
{
\begin{tabular}{lccc}
\hline
\rowcolor[HTML]{EFEFEF} 
\multicolumn{1}{c}{\cellcolor[HTML]{EFEFEF}\textbf{\begin{tabular}[c]{@{}c@{}}Models\\ (K=20)\end{tabular}}} &
  \textbf{\begin{tabular}[c]{@{}c@{}}Aspect\\ Relevance (AR)\end{tabular}} &
  \textbf{\begin{tabular}[c]{@{}c@{}}Sentiment\\ Factuality (SF)\end{tabular}} &
  \textbf{\begin{tabular}[c]{@{}c@{}}Opinion\\ Faithfulness (OF)\end{tabular}} \\ \hline \hline
\textbf{BM25}  &       &       &       \\
\enspace \enspace \ThickVLine \textbf{--} \textbf{\texttt{Mistral-7B}}     & 70.91 & 81.97 & 85.11 \\
\enspace \enspace \ThickVLine \textbf{--} \textbf{\texttt{Llama-3-8B}}     & 80.69 & 89.39 & \textbf{90.37} \\
\enspace \enspace \ThickVLine \textbf{--} \textbf{\texttt{Gemma-2-9B}}     & 78.96 & 87.53 & 84.24 \\
\enspace \enspace \ThickVLine \textbf{--} \textbf{\texttt{Claude-3-Haiku}}    & \textbf{84.87} & 89.99 & 83.48 \\
\enspace \enspace \ThickVLine \textbf{--} \textbf{\texttt{GPT-4o-mini}}    & 78.10  & \textbf{90.45} & 89.53 \\ \hline
\textbf{Dense} &       &       &       \\
\enspace \enspace \ThickVLine \textbf{--} \textbf{\texttt{Mistral-7B}}     & 67.39 & 86.17 & 86.69 \\
\enspace \enspace \ThickVLine \textbf{--} \textbf{\texttt{Llama-3-8B}}     & 75.80 & 91.81 & 89.61 \\
\enspace \enspace \ThickVLine \textbf{--} \textbf{\texttt{Gemma-2-9B}}     & 78.51 & 90.32 & 85.35 \\
\enspace \enspace \ThickVLine \textbf{--} \textbf{\texttt{Claude-3-Haiku}}    & \textbf{81.41} & 91.37 & 83.84 \\
\enspace \enspace \ThickVLine \textbf{--} \textbf{\texttt{GPT-4o-mini}}    & 75.82 & \textbf{92.34} & \textbf{90.03} \\ \hline
\end{tabular}
}
\caption{RAG verification results on the Abstractive summary variant with K=\num{20}, where K is the number of retrieved sentences. Scores are multiplied by \num{100} for better readability.
%and comparison. 
The best results are marked in \textbf{bold}.}
\label{tab:RAG-verification-20}
\end{table}

\paragraph{Results \& Analysis} As presented in Table \ref{tab:RAG-20}, for the Extractive summary variant, the closed-source models (Claude-3-Haiku and GPT-4o-mini) generally outperform the open-source models across all metrics. However, in the Abstractive variant, the performance of open-source models, particularly Llama-3-8B, improves significantly. This suggests that in settings requiring more abstraction and synthesis, open-source models can effectively narrow the gap between themselves and their closed-source counterparts, especially when relevant information is retrieved in a focused manner. In both extractive and abstractive settings, summaries driven by the most important query terms directly impact overall performance. The Oracle baseline further shows that there is still considerable room for improvement, highlighting the inherent challenges in long-form summarization. For RAG verification (Table \ref{tab:RAG-verification-20}), closed-source models outperform open-source models across key metrics. Claude-3-Haiku excels in AR and SF, demonstrating its ability to stay focused on relevant aspects while maintaining factually aligned sentiment. GPT-4o-mini shows strong performance in SF and leads in OF, ensuring that the sentiments and opinions expressed in the generated summaries are consistent with the retrieved evidence. Similar trends are observed with K=\num{10}, as presented in Appendix Tables \ref{tab:RAG-10} and \ref{tab:RAG-verification-10}, which reinforce the results seen with K=\num{20}.

\section{Discussion and Future Directions}
\label{sec:discussions}

\paragraph{Moderation Issues in User Reviews} In the basic setting, Gemini-1.5-Flash generated several invalid summaries due to sensitive or inappropriate content, such as \emph{``Manager is an African middle-aged man who was irresponsible and harsh''} and \emph{``Want more offers?? Call me +1 111 222 ******,''} triggering its safety mechanism\footnote{\href{https://ai.google/responsibility/principles/}{Responsible AI development and AI Principles}}. Even after disabling safety filters, the issue persisted, highlighting the difficulty of handling long-form user reviews. However, in the length-controlled setting, the model produced fewer invalid summaries by prioritizing safer content. Other models did not face similar issues, possibly due to different content moderation filters. Addressing these challenges presents an important area for future work.

\paragraph{Evaluation} Evaluating opinion summarization for long-form user reviews is especially challenging, whether through automatic or human assessments. Human evaluation metrics such as Fluency, Coherence, and Non-Redundancy \cite{brazinskas-etal-2020-shot, angelidis-etal-2021-extractive} are often less applicable when designing systems based on LLMs \cite{song-etal-2024-finesure}. Moreover, most existing LLM-based evaluators are tailored to short input reviews \cite{siledar-etal-2024-one}. Our dataset, with its explicit \texttt{‘PROS’} and \texttt{‘CONS’} paired with long-form reviews, offers opportunities to develop more suitable LLM-based evaluation metrics.
%for long-form opinion summarization.

\section{Conclusion}
\label{sec:conclusion}
In this paper, we addressed key challenges in long-form opinion summarization by introducing a new dataset of over a thousand user reviews per entity, paired with in-depth critical summaries from domain experts. We proposed two training-free summarization methods utilizing RAG and long-context LLMs, designed for scalable and controllable summarization. Additionally, we developed novel reference-free evaluation metrics that offer a fine-grained, context-sensitive assessment of summary faithfulness. Furthermore, based on our insights, we offer suggestions for future research. 
%and development.
%Our benchmarking of both open-source and closed-source LLMs highlights the ongoing challenges of sentiment balance and format adherence, while also showing that open-source models can narrow the gap when relevant information is retrieved effectively.

%mandatory limitations and ethics
\section*{Limitations}
\label{sec:limitations}

In this work, we evaluated our proposed methods using a selection of both open-source and closed-source LLMs. We intentionally focused on cost-effective yet efficient closed-source models and open-source models that can be deployed on consumer-grade hardware, given the constraints of \emph{academic settings}. The performance of more powerful, large-scale models remains unexplored, but we encourage the broader research community to benchmark these models using our dataset and methods. 
%We will make our data, code, and outputs publicly available to support future research. 

While we experimented with different retrievers (BM25 and Dense) for both summary variants using Top-K values of 10 and 20, other retriever configurations might yield better performance. Optimizing for additional retriever options is beyond the scope of this study, but we acknowledge that further exploration in this area could lead to improvements.

Although we proposed novel automatic evaluation metrics built on top of the RAG framework with retrieved evidence, their applicability may be limited in full long-form settings where complete retrieval is not feasible. This remains a potential avenue for future research.

Finally, our research and the development of LFOSum are exclusively centered on the \textbf{English language}. This means its use and effectiveness might not be the same for other languages.

\section*{Ethics Statement}
\label{sec:ethics}

\paragraph{Data Crawling} We carefully considered ethical guidelines when scraping data, ensuring that the data collected is used solely for non-commercial research purposes. Our web scraping was conducted responsibly, at a controlled rate, with the clear intent to avoid any risk of causing a Distributed Denial of Service (\textbf{DDoS}) attack or overloading the servers.

\paragraph{Protection of Privacy} While collecting user reviews, we deliberately chose to exclude any personal information such as reviewer IDs, names, and locations. For our experiments, we focused solely on collecting the review text and date, ensuring that the dataset does not contain any Personally Identifiable Information (\textbf{PII}). This highlights our commitment to user privacy. However, we cannot fully guarantee that users did not include personal details, hate speech, or inappropriate content within the text of their reviews.

% Bibliography entries for the entire Anthology, followed by custom entries
%\bibliography{anthology,custom}
% Custom bibliography entries only
\bibliography{custom}

\clearpage
\appendix
\twocolumn[{%
 \centering
 \Large\bf Supplementary Material: Appendices \\ [20pt]
}]

\begin{table}[t]
\centering
\renewcommand{\arraystretch}{1} % Adjusts the row spacing
\setlength{\tabcolsep}{4pt} % Reduces the space between columns
\resizebox{7.75cm}{!} 
{
\begin{tabular}{lcclc}
\toprule
\rowcolor[HTML]{EFEFEF} 
\multicolumn{2}{c}{\cellcolor[HTML]{EFEFEF}\textbf{TripAdvisor [Source Reviews]}} & \textbf{} & \multicolumn{2}{c}{\cellcolor[HTML]{EFEFEF}\textbf{Oyster [Reference Summary]}} \\ 
\midrule
\#Entities     & 500      &  & \#Entities    & 500    \\
Avg. \#Reviews & $\approx$1.5K &  & Avg. \#Sents  & 12.29  \\
Avg. \#Sents   & $\approx$10.5K    &  & Avg. \#PROS   & 8.45   \\
Avg. \#Words   & $\approx$196K     &  & Avg. \#CONS   & 3.84   \\
Avg. \#Tokens  & $\approx$207K     &  & Avg. \#Tokens & 105.98 \\ 
\bottomrule
\end{tabular}
}
%\caption{Statistics of the LFOSum evaluation dataset. \#Entities denotes the total number of entities. For Source Reviews, averages include user reviews, sentences, words, and tokens (calculated using GPT-4o tokenizer) per entity. For Reference Summary, averages cover sentences, positive/negative sentences, and tokens per entity.}
\caption{Statistics of the LFOSum evaluation dataset. `\#Entities` denotes the total number of entities. For Source Reviews, the averages include the number of user reviews (`Avg. \#Reviews`), sentences (`Avg. \#Sents`), words (`Avg. \#Words`), and tokens (`Avg. \#Tokens`, computed using the GPT-4o tokenizer) per entity. For Reference Summary, the averages represent the number of sentences (`Avg. \#Sents`), positive sentences (`Avg. \#PROS`), negative sentences (`Avg. \#CONS`), and tokens (`Avg. \#Tokens`) per entity.}
\label{tab:dataset-stats}
\end{table}

\vspace{0.5cm}

\begin{table}[t]
\centering
\renewcommand{\arraystretch}{1} % Adjusts the row spacing
\resizebox{7.75cm}{!} 
{
\begin{tabular}{lccc}
\hline
\rowcolor[HTML]{EFEFEF} 
\multicolumn{1}{c}{\cellcolor[HTML]{EFEFEF}\textbf{Models}} & \textbf{PROS Length} & \textbf{CONS Length} & \textbf{Overall} \\ \hline \hline
\textbf{\texttt{GPT-4o-mini}}      & 489 / 500 & 495 / 500 & 484 / 500 \\
\textbf{\texttt{Claude-3-Haiku}}   & 480 / 500 & 455 / 500 & 450 / 500 \\
\textbf{\texttt{Gemini-1.5-Flash}} & 493 / 495 & 490 / 495 & 490 / 495 \\ \hline
\end{tabular}
}
\caption{Length control evaluation results of our Long-form Critic method. ``PROS Length'' refers to the number of generated summaries that adhered to the expected length for positive summaries, while ``CONS Length'' indicates adherence to the length for negative summaries. ``Overall'' represents the total number of summaries where both lengths were followed correctly.}
\label{tab:length-control-results}
\end{table}

\begin{table}[t]
\centering
\renewcommand{\arraystretch}{1} % Adjusts the row spacing
\resizebox{7.75cm}{!} 
{
\begin{tabular}{lccc}
\hline
\rowcolor[HTML]{EFEFEF} 
\multicolumn{1}{c}{\cellcolor[HTML]{EFEFEF}\textbf{Models (K=10)}} & \textbf{R1} & \textbf{RL} & \textbf{BERTScore} \\ \hline \hline
\textbf{Random}       & 16.35          & 8.45         & 47.42         \\
\textbf{Oracle}       & 51.91          & 44.79         & 67.61         \\
\hdashline
\textbf{BM25}               & 20.93      & 10.34     & 53.15     \\
\textbf{Dense}              & 20.58      & 10.56     & 54.26     \\ \hline
\rowcolor[HTML]{EFEFEF} 
\multicolumn{4}{c}{\cellcolor[HTML]{EFEFEF}\textbf{Extractive}}  \\ \hline
\textbf{BM25}               &            &           &           \\
\enspace \enspace \ThickVLine \textbf{--} \textbf{\texttt{Mistral-7B}}         & 20.44 (\textcolor{customRed}{-\num{0.49}})      & 10.26 (\textcolor{customRed}{-\num{0.08}})      & 53.53 (\textcolor{darkgreen}{+\num{0.38}})     \\
\enspace \enspace \ThickVLine \textbf{--} \textbf{\texttt{Llama-3-8B}}         & 21.29 (\textcolor{darkgreen}{+\num{0.36}})      & 10.59 (\textcolor{darkgreen}{+\num{0.25}})      & 54.49 (\textcolor{darkgreen}{+\num{1.34}})     \\
\enspace \enspace \ThickVLine \textbf{--} \textbf{\texttt{Gemma-2-9B}}         & 21.59 (\textcolor{darkgreen}{+\num{0.66}})       & 10.87 (\textcolor{darkgreen}{+\num{0.53}})     & 54.86 (\textcolor{darkgreen}{+\num{1.71}})     \\
\enspace \enspace \ThickVLine \textbf{--} \textbf{\texttt{Claude-3-Haiku}}            & \textbf{21.91} (\textcolor{darkgreen}{+\num{0.98}})         & \textbf{11.01} (\textcolor{darkgreen}{+\num{0.67}})        & \textbf{55.07} (\textcolor{darkgreen}{+\num{1.92}})         \\ 
\enspace \enspace \ThickVLine \textbf{--} \textbf{\texttt{GPT-4o-mini}}            & 21.86 (\textcolor{darkgreen}{+\num{0.93}})          & 10.97 (\textcolor{darkgreen}{+\num{0.63}})         & 55.06 (\textcolor{darkgreen}{+\num{1.91}})         \\ \hline
\textbf{Dense}              &            &           &           \\
\enspace \enspace \ThickVLine \textbf{--} \textbf{\texttt{Mistral-7B}}         & 21.38 (\textcolor{darkgreen}{+\num{0.80}})     & 10.82 (\textcolor{darkgreen}{+\num{0.26}})     & 54.67 (\textcolor{darkgreen}{+\num{0.41}})    \\
\enspace \enspace \ThickVLine \textbf{--} \textbf{\texttt{Llama-3-8B}}         & 22.09 (\textcolor{darkgreen}{+\num{1.51}})      & 10.99 (\textcolor{darkgreen}{+\num{0.43}})     & 55.32 (\textcolor{darkgreen}{+\num{1.06}})    \\
\enspace \enspace \ThickVLine \textbf{--} \textbf{\texttt{Gemma-2-9B}}         & 22.18 (\textcolor{darkgreen}{+\num{1.60}})     & 11.18 (\textcolor{darkgreen}{+\num{0.62}})      & 55.57 (\textcolor{darkgreen}{+\num{1.31}})      \\
\enspace \enspace \ThickVLine \textbf{--} \textbf{\texttt{Claude-3-Haiku}}            & 22.70 (\textcolor{darkgreen}{+\num{2.12}})         & 11.35 (\textcolor{darkgreen}{+\num{0.79}})         & 55.77 (\textcolor{darkgreen}{+\num{1.51}})        \\ 
\enspace \enspace \ThickVLine \textbf{--} \textbf{\texttt{GPT-4o-mini}}            & \textbf{22.78} (\textcolor{darkgreen}{+\num{2.20}})          & \textbf{11.40} (\textcolor{darkgreen}{+\num{0.84}})        & \textbf{55.92} (\textcolor{darkgreen}{+\num{1.66}})        \\ \hline
\rowcolor[HTML]{EFEFEF} 
\multicolumn{4}{c}{\cellcolor[HTML]{EFEFEF}\textbf{Abstractive}} \\ \hline
\textbf{BM25}               &            &           &           \\
\enspace \enspace \ThickVLine \textbf{--} \textbf{\texttt{Mistral-7B}}         & 22.55 (\textcolor{darkgreen}{+\num{1.62}})      & 11.62 (\textcolor{darkgreen}{+\num{1.28}})     & 56.05 (\textcolor{darkgreen}{+\num{2.90}})     \\
\enspace \enspace \ThickVLine \textbf{--} \textbf{\texttt{Llama-3-8B}}         & \textbf{24.94} (\textcolor{darkgreen}{+\num{4.01}})      & \textbf{13.16} (\textcolor{darkgreen}{+\num{2.82}})     & \textbf{56.29} (\textcolor{darkgreen}{+\num{3.14}})     \\
\enspace \enspace \ThickVLine \textbf{--} \textbf{\texttt{Gemma-2-9B}}         & 22.76 (\textcolor{darkgreen}{+\num{1.83}})       & 11.86 (\textcolor{darkgreen}{+\num{1.52}})     & 55.17 (\textcolor{darkgreen}{+\num{2.02}})     \\
\enspace \enspace \ThickVLine \textbf{--} \textbf{\texttt{Claude-3-Haiku}}            & 22.76 (\textcolor{darkgreen}{+\num{1.83}})         & 12.30 (\textcolor{darkgreen}{+\num{1.96}})        & 55.70 (\textcolor{darkgreen}{+\num{2.55}})        \\ 
\enspace \enspace \ThickVLine \textbf{--} \textbf{\texttt{GPT-4o-mini}}            & 24.49 (\textcolor{darkgreen}{+\num{3.56}})          & 12.79 (\textcolor{darkgreen}{+\num{2.45}})        & 55.68 (\textcolor{darkgreen}{+\num{2.53}})        \\ \hline
\textbf{Dense}              &            &           &           \\
\enspace \enspace \ThickVLine \textbf{--} \textbf{\texttt{Mistral-7B}}         & 22.83 (\textcolor{darkgreen}{+\num{2.25}})      & 11.89 (\textcolor{darkgreen}{+\num{1.63}})     & \textbf{56.29} (\textcolor{darkgreen}{+\num{2.03}})     \\
\enspace \enspace \ThickVLine \textbf{--} \textbf{\texttt{Llama-3-8B}}         & 24.19 (\textcolor{darkgreen}{+\num{3.61}})      & \textbf{13.04} (\textcolor{darkgreen}{+\num{2.48}})     & 55.92 (\textcolor{darkgreen}{+\num{1.66}})     \\
\enspace \enspace \ThickVLine \textbf{--} \textbf{\texttt{Gemma-2-9B}}         & 22.48 (\textcolor{darkgreen}{+\num{1.90}})      & 11.93 (\textcolor{darkgreen}{+\num{1.37}})     & 55.18 (\textcolor{darkgreen}{+\num{0.92}})     \\
\enspace \enspace \ThickVLine \textbf{--} \textbf{\texttt{Claude-3-Haiku}}            & 22.66 (\textcolor{darkgreen}{+\num{2.08}})          & 12.41 (\textcolor{darkgreen}{+\num{1.85}})         & 55.58 (\textcolor{darkgreen}{+\num{1.32}})        \\ 
\enspace \enspace \ThickVLine \textbf{--} \textbf{\texttt{GPT-4o-mini}}            & \textbf{24.48} (\textcolor{darkgreen}{+\num{3.90}})          & 13.00 (\textcolor{darkgreen}{+\num{2.44}})        & 55.92 (\textcolor{darkgreen}{+\num{1.66}})         \\ \hline
\end{tabular}
}
\caption{Evaluation results of our LFOSum: RAG Framework with K=\num{10}, where K is the number of retrieved sentences. The best results compared to their respective baseline models are marked in \textbf{bold}, and $\Delta$ gains are shown in round brackets and highlighted in \textcolor{darkgreen}{green} for improvements and \textcolor{customRed}{red} for declines.}
\label{tab:RAG-10}
\end{table}

\begin{table}[H]
\centering
\renewcommand{\arraystretch}{1} % Adjusts the row spacing
\resizebox{7.75cm}{!} 
{
\begin{tabular}{lccc}
\hline
\rowcolor[HTML]{EFEFEF} 
\multicolumn{1}{c}{\cellcolor[HTML]{EFEFEF}\textbf{\begin{tabular}[c]{@{}c@{}}Models\\ (K=10)\end{tabular}}} &
  \textbf{\begin{tabular}[c]{@{}c@{}}Aspect\\ Relevance (AR)\end{tabular}} &
  \textbf{\begin{tabular}[c]{@{}c@{}}Sentiment\\ Factuality (SF)\end{tabular}} &
  \textbf{\begin{tabular}[c]{@{}c@{}}Opinion\\ Faithfulness (OF)\end{tabular}} \\ \hline \hline
\textbf{BM25}  &       &       &       \\
\enspace \enspace \ThickVLine \textbf{--} \textbf{\texttt{Mistral-7B}}     & 70.81 & 82.09 & 82.71 \\
\enspace \enspace \ThickVLine \textbf{--} \textbf{\texttt{Llama-3-8B}}     & 80.97 & 87.54 & \textbf{87.36} \\
\enspace \enspace \ThickVLine \textbf{--} \textbf{\texttt{Gemma-2-9B}}     & 79.18 & 87.58 & 82.09 \\
\enspace \enspace \ThickVLine \textbf{--} \textbf{\texttt{Claude-3-Haiku}}    & \textbf{85.40} & 88.40 & 80.28 \\
\enspace \enspace \ThickVLine \textbf{--} \textbf{\texttt{GPT-4o-mini}}        & 78.10     & \textbf{89.14}    & 87.09     \\ \hline
\textbf{Dense} &       &       &       \\
\enspace \enspace \ThickVLine \textbf{--} \textbf{\texttt{Mistral-7B}}     & 67.74 & 85.54 & 86.53 \\
\enspace \enspace \ThickVLine \textbf{--} \textbf{\texttt{Llama-3-8B}}     & 79.41 & 91.05 & 88.44 \\
\enspace \enspace \ThickVLine \textbf{--} \textbf{\texttt{Gemma-2-9B}}     & 78.58 & 90.59 & 83.82 \\
\enspace \enspace \ThickVLine \textbf{--} \textbf{\texttt{Claude-3-Haiku}}    & \textbf{82.36} & 90.84 & 80.98 \\
\enspace \enspace \ThickVLine \textbf{--} \textbf{\texttt{GPT-4o-mini}}        & 77.72     & \textbf{92.00}     & \textbf{88.92}     \\ \hline
\end{tabular}
}
\caption{RAG verification results on the Abstractive summary variant with K=\num{10}, where K is the number of retrieved sentences. Scores are multiplied by \num{100} for better readability. The best results are marked in \textbf{bold}.}
\label{tab:RAG-verification-10}
\end{table}

\section{Experimental Setup}
\label{sec:experimental-setup}

\subsection{Model Configuration} 

For our RAG Framework, we utilize both open-source models (\texttt{Mistral-7B}, \texttt{Llama-3-8B}, \texttt{Gemma-2-9B}) and closed-source models (\texttt{Claude-3-Haiku}, \texttt{GPT-4o-mini}). Across all models, we set consistent hyperparameters for both the Extractive and Abstractive summarization variants: \texttt{max\_new\_tokens}=256, \texttt{temperature}=0.7, and \texttt{top\_p}=0.9.

For the Long-form Critic, we retain the default parameters of the long-context LLMs (\texttt{GPT-4o-mini}, \texttt{Claude-3-Haiku}, \texttt{Gemini-1.5-Flash}), with the exception of \texttt{max\_tokens}=512, as this value ensures the model can generate comprehensive critic summaries for long-form user reviews.

\subsection{JSON Format Adherence} 

To ensure that the LLMs output in a structured JSON format, we employ several strategies. These include explicitly stating the requirement for JSON output in the prompts, providing a sample JSON structure, and incorporating in-context examples with the desired format. For models such as those from OpenAI\footnote{\href{https://openai.com/index/introducing-structured-outputs-in-the-api/}{Structured Outputs API}, released on August 6th, 2024.} (\texttt{GPT-4o-mini}), we specify formatting instructions by configuring the necessary fields and descriptions (e.g., \texttt{response\_format={``type'': ``json\_object''}}). Similarly, for Gemini models, we use field descriptions (e.g., \texttt{generation\_config={``response\_mime\_type'': ``application/json''}}) to enforce JSON outputs, ensuring reliable evaluation.

\section{Common JSON Parsing Errors} 
\label{sec:parsing-errors}

One of the key challenges when working with LLMs to generate sentiment and length-controlled summaries is ensuring that the outputs conform to a structured format, such as JSON. While the desired output is a well-formed JSON dictionary, LLMs sometimes produce outputs that are incomplete, malformed, or improperly structured, making them difficult or impossible to parse directly. Below, we outline the expected JSON format and common types of issues encountered when generating JSON from LLMs:

\begin{tcolorbox}[colback=lightgray, colframe=mildgray, title=Expected JSON Structure, fonttitle=\bfseries]
\scriptsize
\begin{tabbing}
\hspace{0.5cm}\= \kill % Alignments for nested JSON
\{ \\
\> \textcolor{darkred}{"pros"}: \textcolor{darkgreen}{[} \\
\> \hspace{1cm} \textcolor{darkgreen}{"Central downtown location"}, \\
\> \hspace{1cm} \textcolor{darkgreen}{"Fremont Street Experience next door"}, \\
\> \hspace{1cm} \textcolor{darkgreen}{"Clean and quiet affordable rooms"}, \\
\> \hspace{1cm} \textcolor{darkgreen}{"Four restaurants and bars on-site"}, \\
\> \hspace{1cm} \textcolor{darkgreen}{"Lively casino with penny slots"}, \\
\> \hspace{1cm} \textcolor{darkgreen}{"Access to rooftop pool at California Hotel"}, \\
\> \hspace{1cm} \textcolor{darkgreen}{"On-site parking garage"} \\
\> \textcolor{darkgreen}{]}, \\
\> \textcolor{darkred}{"cons"}: \textcolor{darkgreen}{[} \\
\> \hspace{1cm} \textcolor{darkgreen}{"No easy access to Las Vegas Strip"}, \\
\> \hspace{1cm} \textcolor{darkgreen}{"Noisy common areas"}, \\
\> \hspace{1cm} \textcolor{darkgreen}{"Slight smoke smell throughout hotel"}, \\
\> \hspace{1cm} \textcolor{darkgreen}{"No on-site pool or fitness center"}, \\
\> \hspace{1cm} \textcolor{darkgreen}{"Wi-Fi fee"} \\
\> \textcolor{darkgreen}{]} \\
\}
\end{tabbing}
\end{tcolorbox}

\paragraph{Incomplete Fields:} LLMs generate partial outputs where entire fields, such as 'pros' or 'cons', are missing, incomplete, or malformed.

\begin{tcolorbox}[colback=lightgray, colframe=mildgray, title=Incomplete Fields, fonttitle=\bfseries]
\footnotesize
\begin{tabbing}
\hspace{0.5cm}\= \kill % Alignments for nested JSON
\{ \\
\> \textcolor{darkred}{"pros"}: \textcolor{darkgreen}{[} \\
\> \hspace{1cm} \textcolor{darkgreen}{"Great location"}, \\
\> \hspace{1cm} \textcolor{darkgreen}{"Free Wi-Fi"} \\
\> \textcolor{darkgreen}{]} \\
\> \textcolor{darkred}{"cons"}: \textcolor{darkgreen}{[} \\
\> \hspace{1cm} \textcolor{darkgreen}{"Room was small"}, \\
\> \hspace{1cm} \textcolor{darkgreen}{"Parking is expensive"} \\
%\> \textcolor{darkgreen}{]} \\
\}
\end{tabbing}
\end{tcolorbox}

In this case, the missing comma after the "pros" list and the unclosed string in the "cons" list render this output invalid for parsing.

\paragraph{Incorrect Quotation Marks:} Inconsistent use of single (`) and double (") quotes is a common issue, as JSON requires strict adherence to double quotes for both keys and values.

\begin{tcolorbox}[colback=lightgray, colframe=mildgray, title=Incorrect Quotation Marks, fonttitle=\bfseries]
\footnotesize
\begin{tabbing}
\hspace{0.5cm}\= \kill % Alignments for nested JSON
\{ \\
\> \textcolor{darkred}{`pros'}: \textcolor{darkgreen}{[} \\
\> \hspace{1cm} \textcolor{darkgreen}{`Clean rooms'}, \\
\> \hspace{1cm} \textcolor{darkgreen}{`Good service'} \\
\> \textcolor{darkgreen}{]}, \\
\> \textcolor{darkred}{`cons'}: \textcolor{darkgreen}{[} \\
\> \hspace{1cm} \textcolor{darkgreen}{`No free breakfast'} \\
\> \textcolor{darkgreen}{]} \\
\}
\end{tabbing}
\end{tcolorbox}

This output uses single quotes, making it incompatible with standard JSON parsers.

\paragraph{Extraneous or Missing Commas:} LLMs often omit or misplace commas between key-value pairs or list elements, which breaks the JSON structure.

\begin{tcolorbox}[colback=lightgray, colframe=mildgray, title=Extraneous or Missing Commas, fonttitle=\bfseries]
\footnotesize
\begin{tabbing}
\hspace{0.5cm}\= \kill % Alignments for nested JSON
\{ \\
\> \textcolor{darkred}{"pros"}: \textcolor{darkgreen}{[} \\
\> \hspace{1cm} \textcolor{darkgreen}{"Great location"} \\
\> \hspace{1cm} \textcolor{darkgreen}{"Comfortable beds"} \\
\> \textcolor{darkgreen}{]}, \\
\> \textcolor{darkred}{"cons"}: \textcolor{darkgreen}{[} \\
\> \hspace{1cm} \textcolor{darkgreen}{"No parking"}, , \\
\> \hspace{1cm} \textcolor{darkgreen}{"Room was noisy"} \\
\> \textcolor{darkgreen}{]} \\
\}
\end{tabbing}
\end{tcolorbox}

The missing comma between ``Great location'' and ``Comfortable beds'' and invalid comma between ``No parking'' and ``Room was noisy'', render this JSON invalid.

\paragraph{Mismatched Brackets:} Unbalanced or missing curly braces (\{\}) and square brackets ([]) are frequent, especially when generating long lists or deeply nested structures.

\begin{tcolorbox}[colback=lightgray, colframe=mildgray, title=Mismatched Brackets, fonttitle=\bfseries]
\footnotesize
\begin{tabbing}
\hspace{0.5cm}\= \kill % Alignments for nested JSON
\{ \\
\> \textcolor{darkred}{"pros"}: \textcolor{darkgreen}{[} \\
\> \hspace{1cm} \textcolor{darkgreen}{"Good service"}, \\
\> \hspace{1cm} \textcolor{darkgreen}{"Clean room"} \\
\> \textcolor{darkgreen}{]}, \\
\> \textcolor{darkred}{"cons"}: \textcolor{darkgreen}{[} \\
\> \hspace{1cm} \textcolor{darkgreen}{"Small bathroom"} \\
\> \textcolor{darkgreen}{]} \\
%\}
\end{tabbing}
\end{tcolorbox}

In this case, the closing curly brace is missing, leading to a syntax error.

\paragraph{Output in Bullet Points:} LLM outputs are sometimes structured informally (e.g., using bullet points to list pros and cons), a common format in user-generated content. This structure cannot be directly parsed, as shown in the following example:

\begin{tcolorbox}[colback=lightgray, colframe=mildgray, title=Output in Bullet Points, fonttitle=\bfseries]
\footnotesize
\begin{tabbing}
\hspace{0.5cm}\= \kill % Alignments for nested structure
\textcolor{darkred}{Pros:} \\
\> \textcolor{darkgreen}{- Great location} \\
\> \textcolor{darkgreen}{- Friendly staff} \\
\\
\textcolor{darkred}{Cons:} \\
\> \textcolor{darkgreen}{- Small rooms} \\
\> \textcolor{darkgreen}{- Expensive parking} \\
\end{tabbing}
\end{tcolorbox}

\paragraph{Output in Numbered Lists:} Outputs can also appear in a numbered list format. Due to formatting inconsistencies, these cannot be parsed directly. This issue was particularly observed during our experiments with length-controlled summary generation, as many user reviews present pros and cons in this format.

\begin{tcolorbox}[colback=lightgray, colframe=mildgray, title=Output in Numbered Lists, fonttitle=\bfseries]
\footnotesize
\begin{tabbing}
\hspace{0.5cm}\= \kill % Alignments for nested structure
\textcolor{darkred}{Pros:} \\
\> \textcolor{darkgreen}{1. Clean rooms} \\
\> \textcolor{darkgreen}{2. Friendly staff} \\
\\
\textcolor{darkred}{Cons:} \\
\> \textcolor{darkgreen}{1. No free breakfast} \\
\> \textcolor{darkgreen}{2. Noisy neighbors} \\
\end{tabbing}
\end{tcolorbox}

\paragraph{Minimal Structure:} In some cases, LLM outputs can include lists of pros and cons presented as comma-separated strings within a sentence-like format. This structure often deviates from standard JSON formatting, making it difficult to parse directly, as shown in the following example.

\begin{tcolorbox}[colback=lightgray, colframe=mildgray, title=Minimal Structure, fonttitle=\bfseries]
\footnotesize
\begin{tabbing}
\hspace{0.5cm}\= \kill % Alignments for nested structure
\textcolor{darkred}{Pros:} \textcolor{darkgreen}{"Spacious rooms", "Friendly staff"} \\
\textcolor{darkred}{Cons:} \textcolor{darkgreen}{"No parking", "Small bathroom"} \\
\end{tabbing}
\end{tcolorbox}

\begin{figure}[t]
    \centering
    \includegraphics[scale = 0.62]{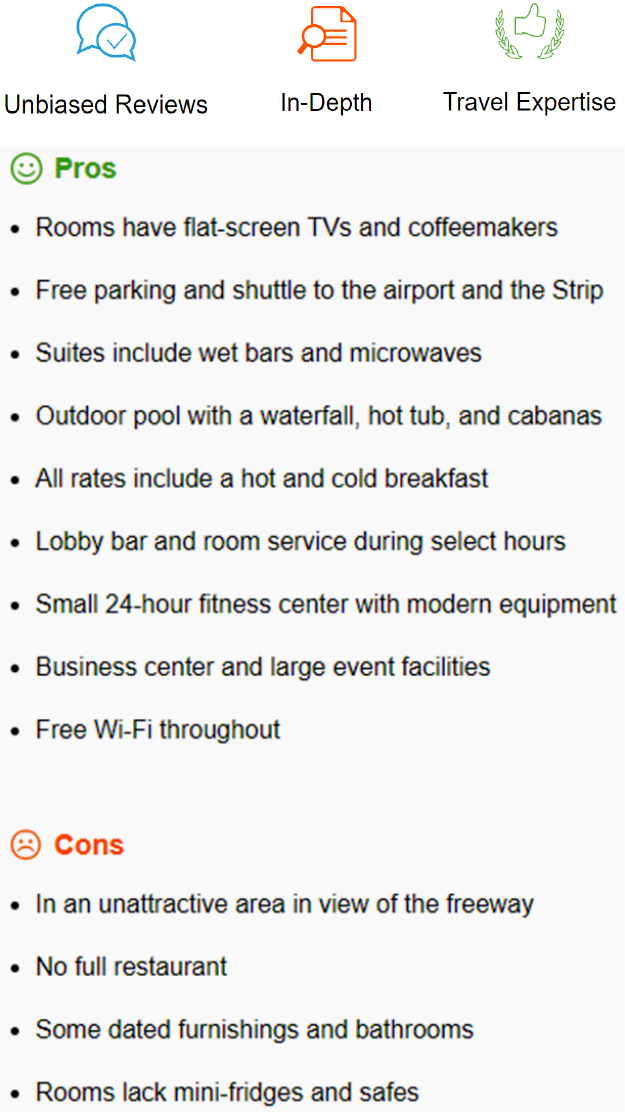}
    \caption{A sample example from our dataset. Hampton Inn Tropicana (\url{https://www.oyster.com/las-vegas/hotels/hampton-inn-tropicana/})}
    \label{fig:sample-example}
\end{figure}

\section{System Message Design}

To guide the LLM for opinion summarization, we developed a system message specifying the model's role and constraints. The message defines the LLM as an “expert summarizer of user reviews” within the domain of “hotels and restaurants,” with a specialization in “travel.” These elements were designed with several key considerations:

\textbf{Role and Task:} Defining the LLM as an expert ensures focused, high-quality outputs. It helps the model capture relevant sentiments and aspects while minimizing irrelevant details.

\textbf{Domain:} Narrowing the scope to hotels and restaurants ensures the model prioritizes key factors such as service quality, location, and amenities—critical in user-generated travel reviews.

\textbf{Specialization:} Adding a travel specialization refines the model’s focus on aspects unique to travelers, such as proximity to attractions and comfort during stays.

\begin{tcolorbox}[colback=lightgray, colframe=mildgray, 
    coltitle=white, fonttitle=\bfseries, title=System Message: RAG Framework, 
    rounded corners, boxrule=0.5mm, width=\linewidth]
\small
\centering You are an \textbf{expert summarizer} of user reviews for \textbf{hotels and restaurants}, specializing in \textbf{travel}!

\end{tcolorbox}

\begin{tcolorbox}[colback=lightgray, colframe=mildgray, 
    coltitle=white, fonttitle=\bfseries, title=System Message: Long-form Critic, 
    rounded corners, boxrule=0.5mm, width=\linewidth]
\small
\centering You are an \textbf{expert critical summarizer} of user reviews for \textbf{hotels and restaurants}, specializing in \textbf{travel}. You provide in-depth evaluations divided into two explicit sections: 'pros' and 'cons', which are reliable summaries.

\end{tcolorbox}

This system messages are crafted to align the model’s outputs with user needs, ensuring summaries remain concise, relevant, and actionable for travel-related decisions.

\section{Data Preprocessing} 
\label{sec:preprocessing}

In our preprocessing pipeline, we focused on filtering the review text based on language without making any explicit modifications to the content of the reviews themselves. We retained only sentences written in English, removing those written in other languages to ensure consistency in the dataset, similar to previous approaches~\citep{nayeem-rafiei-2023-role}. For language identification, we employed the \texttt{spacy-langdetect}\footnote{\url{https://pypi.org/project/spacy-langdetect/}} module, which allowed us to efficiently detect and filter out non-English content, following practices outlined in recent work~\citep{nayeem2024kidlm}.

\section{Related Work}
\label{sec:related-work}

\paragraph{Opinion Summarization Methods} Opinion summarization can generally be divided into two main types: extractive and abstractive. Extractive approaches create summaries by selecting representative sentences directly from the input reviews \citep{angelidis-etal-2021-extractive, basu-roy-chowdhury-etal-2022-unsupervised, li-etal-2023-aspect, chowdhury2024incremental, li-chaturvedi-2024-rationale}. While these methods are scalable and inherently provide traceability to the original content, they often lead to summaries that are overly detailed and lack coherence~\citep{nayeem-chali-2017-extract}. In contrast, abstractive methods generate summaries by synthesizing and rephrasing information from the input reviews \citep{ganesan-etal-2010-opinosis, pmlr-v97-chu19b, brazinskas-etal-2020-unsupervised, amplayo-lapata-2020-unsupervised, hosking2024hierarchical}. This results in summaries that are more fluent and cohesive~\citep{nayeem-etal-2018-abstractive}, though they may require more computational resources and can sometimes lack attribution. Recent advances in LLMs have facilitated the development of opinion summarization models capable of generating effective summaries \cite{bhaskar-etal-2023-prompted} and evaluating the models \cite{siledar-etal-2024-one}, even in zero-shot settings. In this paper, we leverage long-context LLMs to tackle the challenges of long-form opinion summarization, enabling more controllable and scalable summarization techniques tailored to user needs.

\paragraph{Opinion Summarization Datasets} Annotated datasets that pair summaries with reviews are rare, primarily because review platforms do not typically provide summaries, and creating them would require expensive human annotation. To overcome this limitation, previous studies have utilized self-supervised methods to generate synthetic pairs from review corpora \cite{amplayo-lapata-2020-unsupervised, elsahar-etal-2021-self}, where one review is selected as a pseudo-summary and the remaining reviews serve as the input. However, most of these datasets are constrained to a maximum of \num{10} reviews per entity \cite{angelidis-lapata-2018-summarizing, pmlr-v97-chu19b, brazinskas-etal-2020-shot}, with only a few expanding to hundreds \cite{angelidis-etal-2021-extractive, brazinskas-etal-2021-learning}. In reality, many entities accumulate thousands of reviews. A recent effort has aimed to scale opinion summarization \cite{muddu2024distill}, but their dataset, annotated using GPT-4 rather than human annotators, focuses on product reviews (see \sref{sec:dataset} and Table \ref{tab:dataset-comparison} for a discussion on the scarcity of long-form input documents in product reviews) and lacks true book-length inputs (> \num{100}K tokens) \cite{chang2024booookscore}\footnote{While the dataset is not publicly available, Table 2 in the paper suggests an average of approximately 61,411 words per entity.}. In this paper, we introduce a new dataset of long-form user reviews, each entity featuring over a thousand reviews, paired with in-depth and unbiased critical summaries provided by domain experts. This dataset offers fresh opportunities for evaluating and analyzing the capabilities of opinion summarization models, especially when managing large-scale, diverse inputs that resemble book-length documents.

\begin{figure*}[t]
    \centering
    \includegraphics[scale = 0.65]{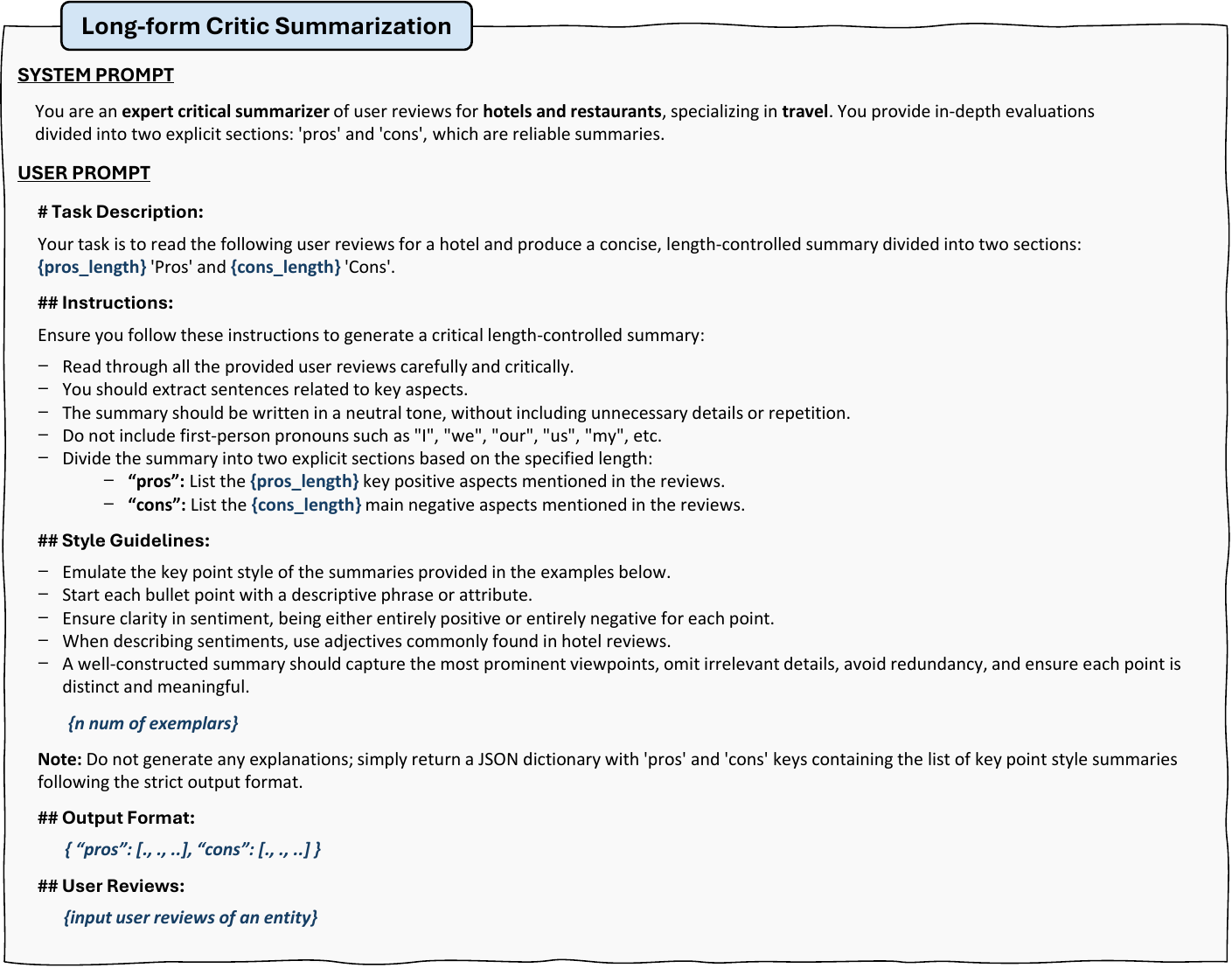}
    \caption{Long-form Critic Summarization Prompt.}
    \label{fig:critic-prompt}
\end{figure*}

\begin{figure*}[t]
    \centering
    \includegraphics[scale = 0.65]{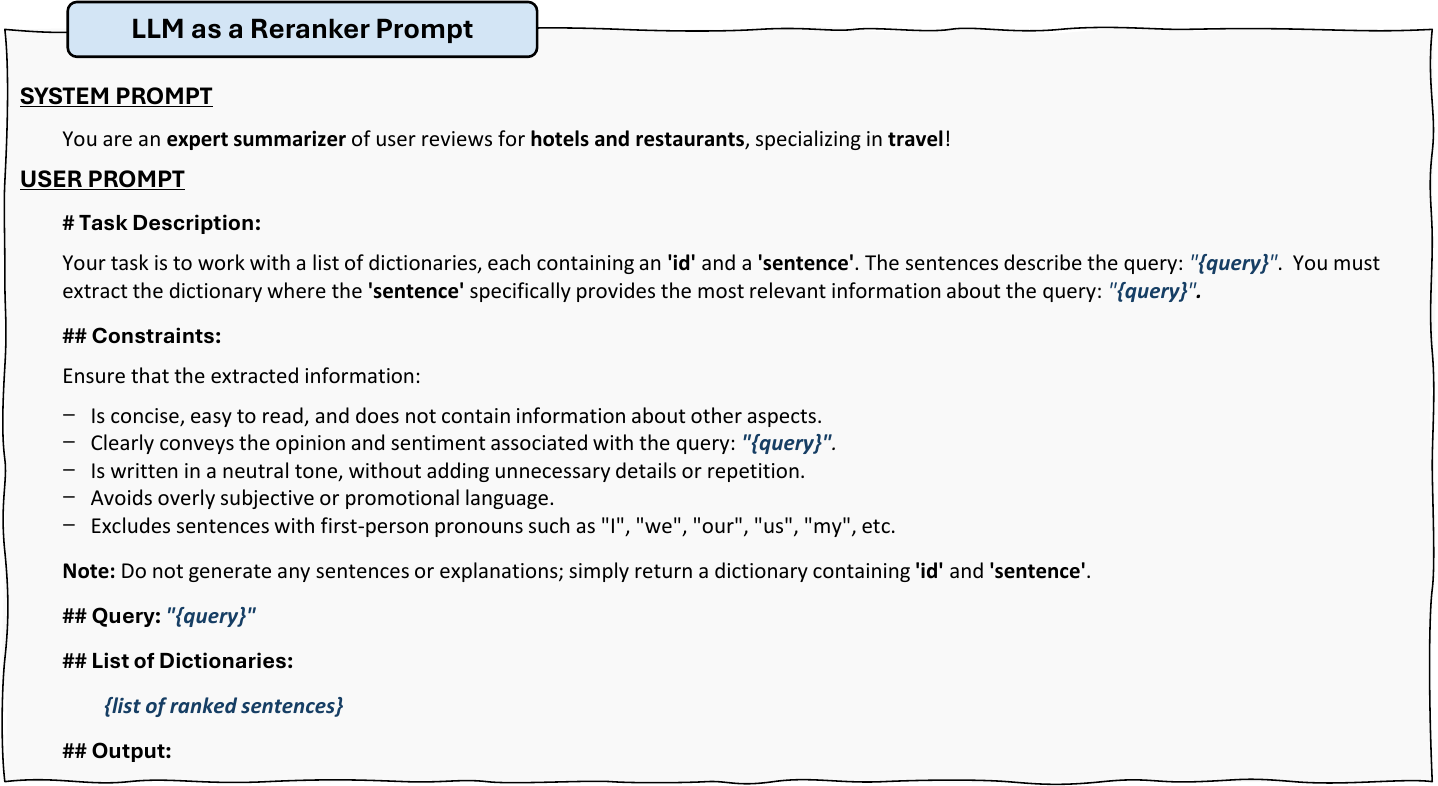}
    \caption{LLM as a Reranker Prompt.}
    \label{fig:reranker-prompt}
\end{figure*}

\begin{figure*}[t]
    \centering
    \includegraphics[scale = 0.65]{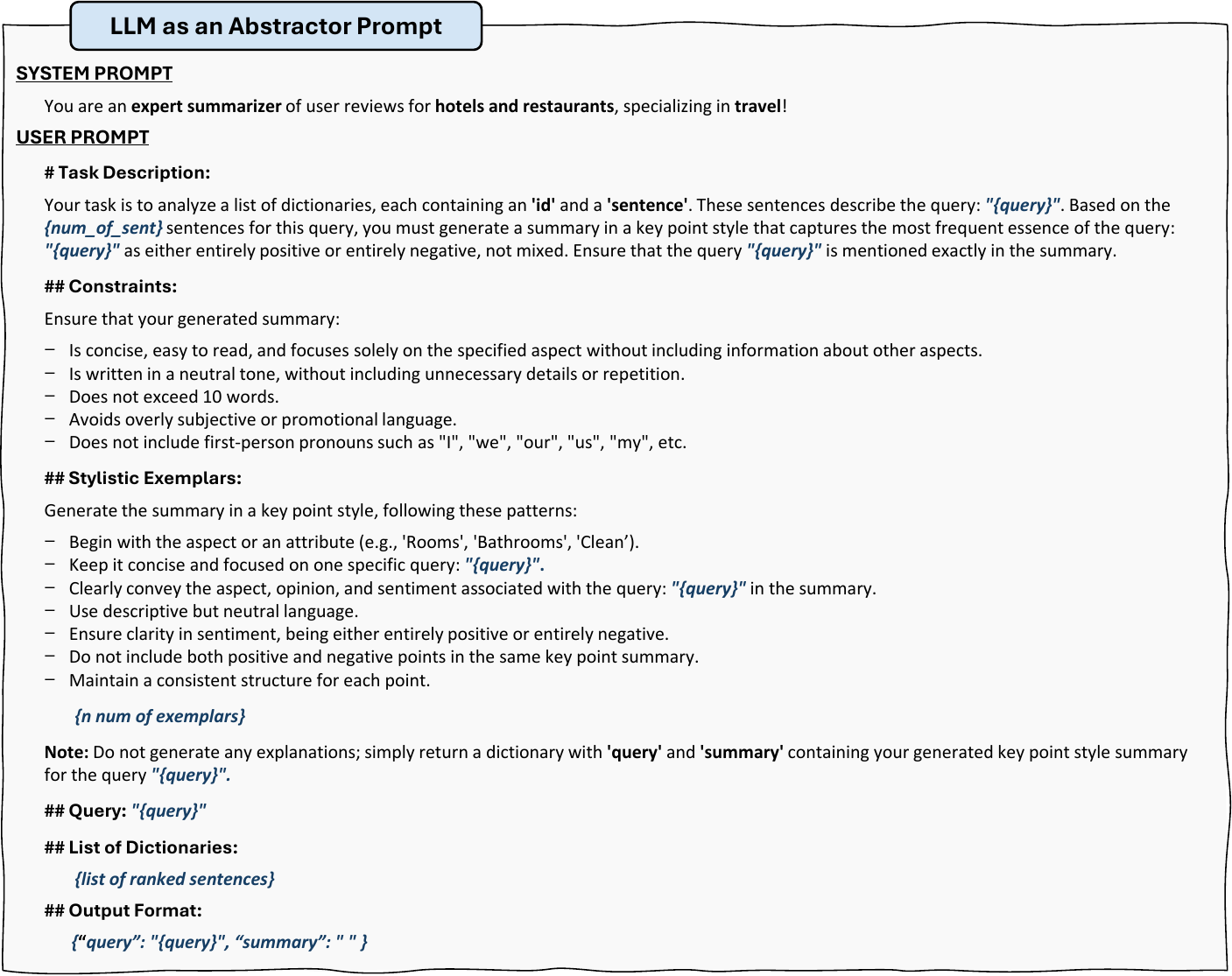}
    \caption{LLM as an Abstractor Prompt.}
    \label{fig:abstractor-prompt}
\end{figure*}

%% Example Critic Summaries#1
\begin{table*}[t]
\centering
\caption{\textbf{[Example\#1]} - Summaries generated by different LLMs in our Long-form Critic model with length control settings. The reference summary is {\ul underlined}, and the `Pros' and `Cons' are highlighted in  \textcolor{darkgreen}{green} and \textcolor{customRed}{red}, respectively.}
\scriptsize % Smaller font size
\begin{adjustbox}{max width=\textwidth}
\begin{tabularx}{\textwidth}{|>{\raggedright\arraybackslash}p{2.5cm}|>{\raggedright\arraybackslash}p{6.25cm}|>{\raggedright\arraybackslash}p{6cm}|}
\hline
\rowcolor{lightgray} \textbf{Summary} & \textbf{Pros} & \textbf{Cons} \\
\hline
\textbf{{\ul Reference Summary}} & 
\vspace{0.2cm} % Adds space before
-- \textcolor{darkgreen}{Budget accommodations on the Upper West Side.} \newline
-- \textcolor{darkgreen}{An attractive and somewhat peaceful NYC residential neighborhood.} \newline
-- \textcolor{darkgreen}{Across the street from Riverside Park; walking distance to Central Park.} \newline
-- \textcolor{darkgreen}{Basic rooms include microwaves and mini-fridges.} \newline
-- \textcolor{darkgreen}{Private bathrooms (though small), and air-conditioning units.} \newline
-- \textcolor{darkgreen}{Updates such as new tiling and flat-screen televisions.} \newline
-- \textcolor{darkgreen}{Higher floors may have views of the Hudson River.} \newline
-- \textcolor{darkgreen}{Three blocks from the subway (the 1 train).} \newline
-- \textcolor{darkgreen}{Suites have two separate bedrooms for privacy.} \newline
-- \textcolor{darkgreen}{Those arriving by car can usually find street parking (if patient).} \newline
-- \textcolor{darkgreen}{Free Wi-Fi.} 
\vspace{0.2cm} % Adds space after
& 
\vspace{0.2cm} % Adds space before
-- \textcolor{darkred}{Rooms have spartan furnishings, and dated comforters are an eyesore.} \newline
-- \textcolor{darkred}{Though clean on the surface, the hotel shows its age.} \newline
-- \textcolor{darkred}{Not all rooms have been renovated with new tiling and televisions.} \newline
-- \textcolor{darkred}{Bathrooms are tight on space, especially the walk-in showers.} \newline
-- \textcolor{darkred}{Not all rooms have work desks.} \newline
-- \textcolor{darkred}{The older compact elevator may disturb claustrophobic guests.} 
\vspace{0.2cm} % Adds space after
\\
\hline

\textbf{GPT-4o-mini} & 
\vspace{0.2cm} % Adds space before
-- \textcolor{darkgreen}{Excellent location near Riverside Park and Central Park.} \newline
-- \textcolor{darkgreen}{Friendly and helpful staff, often going above and beyond.} \newline
-- \textcolor{darkgreen}{Affordable rates compared to other Manhattan hotels.} \newline
-- \textcolor{darkgreen}{Rooms are clean with daily housekeeping service.} \newline
-- \textcolor{darkgreen}{Free Wi-Fi available in the lobby.} \newline
-- \textcolor{darkgreen}{Close to multiple subway stations.} \newline
-- \textcolor{darkgreen}{Views of the Hudson River from upper-level rooms.} \newline
-- \textcolor{darkgreen}{Microwave and refrigerator included in rooms.} \newline
-- \textcolor{darkgreen}{Safe and quiet neighborhood for nighttime walks.} \newline
-- \textcolor{darkgreen}{Diverse dining options and cafés nearby, including Zabar's.} \newline
-- \textcolor{darkgreen}{Functional space for budget travelers not seeking luxury.}
\vspace{0.2cm} % Adds space after
& 
\vspace{0.2cm} % Adds space before
-- \textcolor{darkred}{Rooms are very small, often cramped for multiple guests.} \newline
-- \textcolor{darkred}{Minimal amenities with no in-room toiletries provided.} \newline
-- \textcolor{darkred}{Older building showing signs of wear and tear.} \newline
-- \textcolor{darkred}{Tiny and slow elevators, causing inconvenience.} \newline
-- \textcolor{darkred}{Occasional cleanliness issues in common areas.} \newline
-- \textcolor{darkred}{Noise from plumbing and neighboring rooms can be bothersome.} 
\vspace{0.2cm} % Adds space after
\\
\hline

%\vspace{0.5cm}  % Adding vertical space between sections
\textbf{Claude-3-Haiku} & 
\vspace{0.2cm} % Adds space before
-- \textcolor{darkgreen}{Excellent location on the Upper West Side, close to Central Park, subway, restaurants, and shops.} \newline
-- \textcolor{darkgreen}{Affordable prices, especially for the area.} \newline
-- \textcolor{darkgreen}{Friendly and helpful staff.} \newline
-- \textcolor{darkgreen}{Rooms have mini-fridge and microwave.} \newline
-- \textcolor{darkgreen}{Free WiFi in the lobby.} \newline
-- \textcolor{darkgreen}{Quiet, safe residential neighborhood.} \newline
-- \textcolor{darkgreen}{Good value for the price.} \newline
-- \textcolor{darkgreen}{Rooms are cleaned daily and have private bathrooms.} \newline
-- \textcolor{darkgreen}{Some rooms offer nice views of the Hudson River.} \newline
-- \textcolor{darkgreen}{Convenient to public transportation.} \newline
-- \textcolor{darkgreen}{Suitable for budget-conscious travelers.} 
\vspace{0.2cm} % Adds space after
& 
\vspace{0.2cm} % Adds space before
-- \textcolor{darkred}{Rooms are very small and can feel cramped.} \newline
-- \textcolor{darkred}{Decor and furnishings are dated and worn.} \newline
-- \textcolor{darkred}{Elevators are tiny and slow.} \newline
-- \textcolor{darkred}{Carpets and hallways appear dirty and unkempt.} \newline
-- \textcolor{darkred}{Limited amenities (no breakfast, no cable TV).} 
\vspace{0.2cm} % Adds space after
\\
\hline

%\vspace{0.5cm}  % Adding vertical space between sections
\textbf{Gemini-1.5-Flash} & 
\vspace{0.2cm} % Adds space before
-- \textcolor{darkgreen}{Excellent location near Riverside Park and Central Park.} \newline
-- \textcolor{darkgreen}{Friendly and helpful staff, often going above and beyond.} \newline
-- \textcolor{darkgreen}{Affordable rates compared to other Manhattan hotels.} \newline
-- \textcolor{darkgreen}{Rooms are clean with daily housekeeping service.} \newline
-- \textcolor{darkgreen}{Free Wi-Fi available in the lobby.} \newline
-- \textcolor{darkgreen}{Close to multiple subway stations.} \newline
-- \textcolor{darkgreen}{Views of the Hudson River from upper-level rooms.} \newline
-- \textcolor{darkgreen}{Microwave and refrigerator included in rooms.} \newline
-- \textcolor{darkgreen}{Safe and quiet neighborhood for nighttime walks.} \newline
-- \textcolor{darkgreen}{Diverse dining options and cafés nearby, including Zabar's.} \newline
-- \textcolor{darkgreen}{Functional space for budget travelers not seeking luxury.}
\vspace{0.2cm} % Adds space after
& 
\vspace{0.2cm} % Adds space before
-- \textcolor{darkred}{Rooms are very small, often cramped for multiple guests.} \newline
-- \textcolor{darkred}{Minimal amenities with no in-room toiletries provided.} \newline
-- \textcolor{darkred}{Older building showing signs of wear and tear.} \newline
-- \textcolor{darkred}{Tiny and slow elevators, causing inconvenience.} \newline
-- \textcolor{darkred}{Occasional cleanliness issues in common areas.} \newline
-- \textcolor{darkred}{Noise from plumbing and neighboring rooms can be bothersome.} 
\vspace{0.2cm} % Adds space after
\\
\hline
\end{tabularx}
\end{adjustbox}
\label{tab:output-summaries-critic-1}
\end{table*}

%% Example Critic Summaries#2
\begin{table*}[t]
\centering
\caption{\textbf{[Example\#2]} - Summaries generated by LLMs in our Long-form Critic model with length control settings. The reference summary is {\ul underlined}, and the `Pros' and `Cons' are highlighted in  \textcolor{darkgreen}{green} and \textcolor{customRed}{red}, respectively.}
\scriptsize % Smaller font size
\begin{adjustbox}{max width=\textwidth}
\begin{tabularx}{\textwidth}{|>{\raggedright\arraybackslash}p{2.5cm}|>{\raggedright\arraybackslash}X|>{\raggedright\arraybackslash}X|}
\hline
\rowcolor{lightgray} \textbf{Summary} & \textbf{Pros} & \textbf{Cons} \\
\hline
\textbf{{\ul Reference Summary}} & 
\begin{itemize}
    \item \textcolor{darkgreen}{Rich history}
    \item \textcolor{darkgreen}{Charming Beaux Arts architecture}
    \item \textcolor{darkgreen}{Easily accessible location in the Flatiron District}
    \item \textcolor{darkgreen}{Rooms feature Frette linens, C.O. Bigelow toiletries, and flat-screen TVs}
    \item \textcolor{darkgreen}{Free Wi-Fi}
\end{itemize} & 
\begin{itemize}
    \item \textcolor{darkred}{Ongoing renovations make for a noisy stay}
    \item \textcolor{darkred}{No on-site fitness center, restaurant, or bar}
    \item \textcolor{darkred}{Dated furnishings in some rooms}
    \item \textcolor{darkred}{Limited tourist attractions within walking distance (but subways nearby))}
\end{itemize} \\
\hline

\textbf{GPT-4o-mini} & 
\begin{itemize}
    \item \textcolor{darkgreen}{Friendly and accommodating staff provide exceptional customer service}
    \item \textcolor{darkgreen}{Stylish Art Deco decor creates a warm, welcoming atmosphere}
    \item \textcolor{darkgreen}{Convenient location near subway, parks, and major attractions}
    \item \textcolor{darkgreen}{Complimentary wine hour and delicious pastries from the in-house bakery}
    \item \textcolor{darkgreen}{Well-appointed, clean rooms with comfortable beds and heated bathroom floors}
\end{itemize} & 
\begin{itemize}
    \item \textcolor{darkred}{Rooms can be small and lack adequate storage space}
    \item \textcolor{darkred}{Some noise from neighboring rooms or hallways can be disruptive}
    \item \textcolor{darkred}{Limited breakfast options and lack of in-room coffee facilities}
    \item \textcolor{darkred}{Mandatory facility fee for amenities like Wi-Fi and gym access}
\end{itemize} \\
\hline

\textbf{Claude-3-Haiku} & 
\begin{itemize}
    \item \textcolor{darkgreen}{Excellent location near major attractions, subway lines, and restaurants}
    \item \textcolor{darkgreen}{Stylish, art deco-inspired decor with a cozy, relaxed vibe}
    \item \textcolor{darkgreen}{Friendly and accommodating staff, including helpful concierge services}
    \item \textcolor{darkgreen}{Comfortable beds and well-equipped bathrooms with luxurious amenities}
    \item \textcolor{darkgreen}{Convenient on-site dining options, including a bakery, cafe, and renowned restaurant}
\end{itemize} & 
\begin{itemize}
    \item \textcolor{darkred}{Rooms can be small, even for New York City standards}
    \item \textcolor{darkred}{Potential for noise from the hallway and neighboring rooms}
    \item \textcolor{darkred}{Limited room amenities like coffee makers and mini-fridges}
    \item \textcolor{darkred}{Daily resort fee despite lacking typical resort-style facilities}
\end{itemize} \\
\hline

\textbf{Gemini-1.5-Flash} & 
\begin{itemize}
    \item \textcolor{darkgreen}{Warm and welcoming staff, with personalized service and attention to detail}
    \item \textcolor{darkgreen}{Beautifully decorated rooms with a chic and comfortable aesthetic}
    \item \textcolor{darkgreen}{Convenient location near major attractions, restaurants, and transportation}
    \item \textcolor{darkgreen}{Delicious food and beverage options, including a bakery, casual restaurant, and fine dining}
    \item \textcolor{darkgreen}{Excellent value for money compared to other hotels in NYC}
\end{itemize} & 
\begin{itemize}
    \item \textcolor{darkred}{Rooms can be small, especially for larger groups or long stays}
    \item \textcolor{darkred}{Street noise can be an issue, especially for light sleepers}
    \item \textcolor{darkred}{Limited closet space in some rooms}
    \item \textcolor{darkred}{No coffee makers in rooms, only complimentary coffee and pastry vouchers}
\end{itemize} \\
\hline
\end{tabularx}
\end{adjustbox}
\label{tab:output-summaries-critic-2}
\end{table*}

%% Example Summary RAG
\begin{table*}[ht]
\centering
\caption{\textbf{Summaries generated by different LLMs in Extractive (Dense) and Abstractive (BM25) settings for opinion summarization for TopK (where K=\num{20}) retrieved sentences.}}
\footnotesize
\begin{tabular}{|p{3cm}|p{12cm}|}
\hline
\multicolumn{2}{|c|}{\textbf{Extractive - Dense}} \\
\hline \hline
\vspace{0.3cm}
\textbf{Reference Summary} & \vspace{0.1cm}Great Marina District location surrounded by shops and restaurants. The waterfront is a 12-minute walk away. Free garage and limited street parking. Clean rooms with large windows and flat-screen TVs. Priced well for clean hotel in Marina District City tour bus stops outside, and tickets are available at reception. Free Wi-Fi. Rooms are dated and some get street noise. No scene or lounge/restaurant area. Wi-Fi can be spotty. \vspace{0.1cm}\\
\hline
\vspace{0.3cm}
\textbf{GPT-4o-mini} & \vspace{0.1cm}The room is spacious and gives great privacy. Nice, clean, comfortable hotel with fairly large rooms and free parking. The parking is good and available. The traffic on this street never stops which we did not realise when we booked. But it is a pleasant walk and nice to see the city. Location in Marina/Cow Hollow district on Lombard. You will hear people talking through the night outside. Excellent food and service. The staff were helpful and efficient. The bed was comfortable. Great area for restaurants and walking. Very attractive price for San Francisco which can be ridiculously expensive sometimes. If you have a car, this place is exceptional. Room was well cleaned each day. It is in a clean and seemingly safe part of the city. \vspace{0.1cm}\\
\hline
\vspace{0.3cm}
\textbf{Llama-3-8B} & \vspace{0.1cm}The room is well appointed. Nice, clean, comfortable hotel with fairly large rooms and free parking. Covered parking provided for approx. This property was noisy from the street. But it is a pleasant walk and nice to see the city. Location in Marina/Cow Hollow district right on Lombard. You will hear people talking through the night outside. Good restaurants, Indian, Thai, Chinese, American diner. Helpful knowledgeable staff, good beds. Very busy area. Price, we paid \$167 per night but this was peak season. Our car was broken into. Room was well cleaned each day. Plan of the city at disposal in the lobby. \vspace{0.1cm}\\
\hline \hline
\multicolumn{2}{|c|}{\textbf{Abstractive - BM25}} \\
\hline \hline
\vspace{0.3cm}
\textbf{Reference Summary} & \vspace{0.1cm}Friendly staff. Nice pool and beautifully landscaped pool area. Rooms include free Internet, flat-screen TVs, and iPod docks. Convenient location. Free Wi-Fi in lobby and pool area. Rooms not in great condition. No room service; no poolside drink service. Two blocks from beach, surrounded by motels in slightly grubby location. Lumpy beds. No restaurant on-site. \vspace{0.1cm}\\
\hline
\vspace{0.3cm}
\textbf{GPT-4o-mini} & \vspace{0.1cm} Rooms are spacious and generally well-maintained. Convenient access to the beach and amenities provided. Noisy at night. Parking costs \$20 to \$25 per day. Excellent location with great accessibility and convenience. Hotel staff are consistently helpful and friendly. Everything is within walking distance. Pool area is wonderful and well-maintained. Beds are very comfortable and spacious. Conveniently located just blocks from the beach. Multiple floors with varied room sizes and views. Street noise is significant and bothersome. Welcoming and beautiful lobby atmosphere. Price is great and very reasonable. Service is consistently excellent and highly praised. \vspace{0.1cm}\\
\hline
\vspace{0.3cm}
\textbf{Llama-3-8B} & \vspace{0.1cm} Rooms are spacious and well-appointed. Close proximity to beach with beach towels and chairs provided. Noisy at night. Every day is well-planned and busy. Convenient and perfect location. Extremely helpful and friendly. Walking distance to everything. Pool area is wonderful. Beds are very comfortable and soft. Conveniently located within a few blocks from beach and attractions. The hotel has multiple floors with varying room layouts and ocean views. Busy and noisy. Well-designed and welcoming lobby area. Reasonably priced with great value. Service is exceptional and faultless. \vspace{0.1cm} \\
\hline
\end{tabular}
\label{tab:output-summaries-RAG}
\end{table*}

\end{document}